\PassOptionsToPackage{table}{xcolor}
\documentclass[letterpaper, 10 pt, conference]{ieeeconf}  

\IEEEoverridecommandlockouts 
\title{\LARGE \bf
SLNet: A Super-Lightweight Geometry-Adaptive Network for 3D Point Cloud Recognition
}

\author{
	\parbox{\textwidth}{%
		\centering
		Mohammad Saeid$^{1}$, Amir Salarpour$^{2}$, Pedram MohajerAnsari$^{2}$, Mert D. Pes\'{e}$^{2}$%
	}%
	\thanks{$^{1}$Mohammad Saeid is with Sirjan University of Technology, Sirjan, Iran
		{\tt m.saeid@stu.sirjantech.ac.ir}}%
	\thanks{$^{2}$Amir Salarpour, Pedram MohajerAnsari, and Mert D. Pes\'{e} are with Clemson University, Clemson, SC, USA
		\{\texttt{asalarp}, \texttt{pmohaje}, \texttt{mpese}\}@clemson.edu}
}

\usepackage{graphicx}
\usepackage{amsmath,amssymb,amsfonts}
\usepackage{booktabs}
\usepackage{multirow}
\usepackage{siunitx}
\usepackage{cite}
\usepackage{xspace}

\usepackage{algorithmic}
\usepackage{textcomp}
\usepackage{xcolor}
\usepackage{float}
\usepackage{wrapfig}

\usepackage{colortbl}

\usepackage{array}
\usepackage{makecell}
\usepackage{cellspace}
\usepackage{adjustbox}
\usepackage{threeparttable}
\usepackage{tabularx}

\usepackage{caption}
\usepackage{subcaption}

\usepackage{tikz}
\usetikzlibrary{positioning,backgrounds,shapes.geometric,calc,fit}

\usepackage{pgfplots}
\usepackage{pgfplotstable}
\pgfplotsset{compat=1.18}
\usepgfplotslibrary{polar}

\usepackage[normalem]{ulem}

\newcommand{\SLNet}{\textit{\textbf{SLNet}}\xspace}
\newcommand{\SLNetS}{\textit{\textbf{SLNet\--S}}\xspace}
\newcommand{\SLNetM}{\textit{\textbf{SLNet\--M}}\xspace}
\newcommand{\SLNetT}{\textit{\textbf{SLNet\--T}}\xspace}

\definecolor{lightgreen}{rgb}{0.9,1,0.9}
\def\BibTeX{{\rm B\kern-.05em{\sc i\kern-.025em b}\kern-.08em
    T\kern-.1667em\lower.7ex\hbox{E}\kern-.125emX}}

\usepackage[colorlinks=true, linkcolor=blue, citecolor=green, urlcolor=red]{hyperref}

\usepackage[capitalize]{cleveref}
\crefname{section}{Sec.}{Secs.}
\Crefname{section}{Section}{Sections}
\Crefname{table}{Table}{Tables}
\crefname{table}{Tab.}{Tabs.}
\crefname{figure}{Figure}{Figures}

\begin{document}

\maketitle
\thispagestyle{empty}
\pagestyle{empty}

\begin{abstract}
We present \textbf{SLNet}, a lightweight backbone for 3D point cloud recognition designed to achieve strong performance without the computational cost of many recent attention, graph, and deep MLP based models. The model is built on two simple ideas: \textbf{NAPE} (Nonparametric Adaptive Point Embedding), which captures spatial structure using a combination of Gaussian RBF and cosine bases with input adaptive bandwidth and blending, and \textbf{GMU} (Geometric Modulation Unit), a per channel affine modulator that adds only $2D$ learnable parameters. These components are used within a four stage hierarchical encoder with FPS+kNN grouping, nonparametric normalization, and shared residual MLPs. In experiments, SLNet shows that a very small model can still remain highly competitive across several 3D recognition tasks. On ModelNet40, \textbf{SLNet-S} with 0.14M parameters and 0.31\,GFLOPs achieves 93.64\% overall accuracy, outperforming PointMLP-elite with $5\times$ fewer parameters, while \textbf{SLNet-M} with 0.55M parameters and 1.22\,GFLOPs reaches 93.92\%, exceeding PointMLP with $24\times$ fewer parameters. On ScanObjectNN, SLNet-M achieves 84.25\% overall accuracy within 1.2 percentage points of PointMLP while using $28\times$ fewer parameters. For large scale scene segmentation, \textbf{SLNet-T} extends the backbone with local Point Transformer attention and reaches 58.2\% mIoU on S3DIS Area~5 with only 2.5M parameters, more than $17\times$ fewer than Point Transformer V3. We also introduce \textbf{NetScore$^{+}$}, which extends NetScore by incorporating latency and peak memory so that efficiency can be evaluated in a more deployment oriented way. Across multiple benchmarks and hardware settings, SLNet delivers a strong overall balance between accuracy and efficiency. Code is available at: \url{https://github.com/m-saeid/SLNet}.

\end{abstract}

\section{Introduction}
\label{sec:intro}

Real-time 3D perception is important in applications such as autonomous driving~\cite{li2020deep}, robotics~\cite{duan2021robotics}, and augmented reality~\cite{mahmood2020bim}. In many of these settings, especially on edge devices, models must operate under strict limits on latency, memory, and power. Point clouds remain a widely used 3D representation because they preserve fine geometric detail without quantization artifacts~\cite{dumic2025three}. However, many strong point cloud backbones remain too expensive for resource-constrained deployment, often requiring more than 0.7M parameters and 1\,GFLOP even at 1k input points.

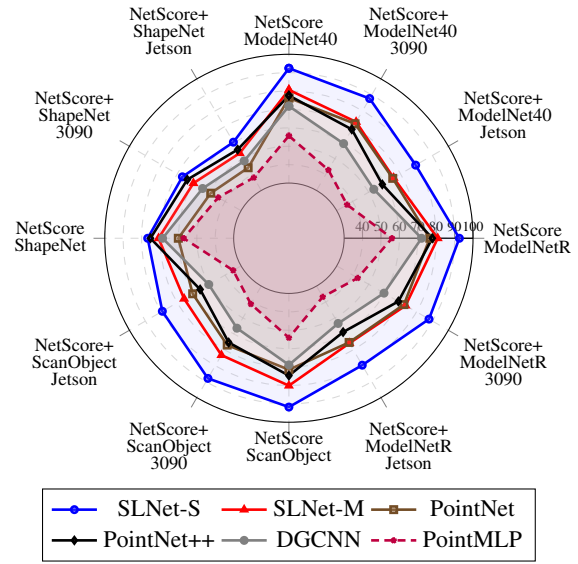
\begin{figure}[t]
\centering
\begin{tikzpicture}
\begin{polaraxis}[
width=0.75\columnwidth,
height=0.75\columnwidth,
ymin=30, ymax=100,
ytick={40,50,60,70,80,90,100},
yticklabels={40,50,60,70,80,90,100},
yticklabel style={font=\scriptsize, opacity=1, scale=0.75},
start angle=90,
xtick={90,60,30,0,330,300,270,240,210,180,150,120},
xticklabel style={font=\scriptsize, align=center, text width=1.2cm},
xticklabels={
{NetScore\\[-0.5ex]ModelNet40\\[-0.5ex]},
{NetScore+\\[-0.5ex]ModelNet40\\[-0.5ex]3090},
{NetScore+\\[-0.5ex]ModelNet40\\[-0.5ex]Jetson},
{NetScore\\[-0.5ex]ModelNetR\\[-0.5ex]},
{NetScore+\\[-0.5ex]ModelNetR\\[-0.5ex]3090},
{NetScore+\\[-0.5ex]ModelNetR\\[-0.5ex]Jetson},
{NetScore\\[-0.5ex]ScanObject\\[-0.5ex]},
{NetScore+\\[-0.5ex]ScanObject\\[-0.5ex]3090},
{NetScore+\\[-0.5ex]ScanObject\\[-0.5ex]Jetson},
{NetScore\\[-0.5ex]ShapeNet\\[-0.5ex]},
{NetScore+\\[-0.5ex]ShapeNet\\[-0.5ex]3090},
{NetScore+\\[-0.5ex]ShapeNet\\[-0.5ex]Jetson}
},
grid=both,
major grid style={dashed, opacity=0.6},
minor grid style={dashed, opacity=0.2},
label distance=5pt,
clip=false,
legend style={at={(0.5,-0.18)}, anchor=north, legend columns=3, font=\small},
]
\addplot+[sharp plot, every mark/.append style={fill=black}, mark=*, mark options={scale=0.6}, fill=blue!30, fill opacity=0.15, line width=1pt]
coordinates {
(90,92.42) (60,87.71) (30,79.50)
(0,92.59) (330,87.87) (300,79.66)
(270,91.76) (240,87.96) (210,79.38)
(180,76.70) (150,66.81) (120,60.31)
(90,92.42)
};
\addplot+[sharp plot, mark=triangle*, mark options={scale=0.6}, fill=teal!30, fill opacity=0.15, line width=1pt]
coordinates {
(90,80.66) (60,73.01) (30,65.67)
(0,80.83) (330,73.18) (300,65.83)
(270,80.12) (240,73.34) (210,65.73)
(180,70.81) (150,59.88) (120,53.43)
(90,80.66)
};
\addplot+[sharp plot, mark=square*, mark options={scale=0.6}, fill=orange!30, fill opacity=0.15, line width=1pt]
coordinates {
(90,76.27) (60,71.86) (30,65.13)
(0,76.53) (330,72.12) (300,65.39)
(270,71.45) (240,67.04) (210,60.35)
(180,60.07) (150,49.03) (120,44.27)
(90,76.27)
};
\addplot+[sharp plot, mark=diamond*, mark options={scale=0.6}, fill=red!30, fill opacity=0.12, line width=1pt]
coordinates {
(90,77.59) (60,68.34) (30,58.57)
(0,77.91) (330,68.66) (300,58.89)
(270,74.66) (240,65.42) (210,55.66)
(180,75.19) (150,63.74) (120,55.66)
(90,77.59)
};
\addplot+[sharp plot, mark=otimes*, mark options={scale=0.6}, fill=purple!30, fill opacity=0.12, line width=1pt, color=gray]
coordinates {
(90,71.84) (60,59.42) (30,53.25)
(0,72.06) (330,59.65) (300,53.47)
(270,68.85) (240,56.44) (210,50.27)
(180,68.62) (150,54.13) (120,48.56)
(90,71.84)
};
\addplot+[sharp plot, mark=star, mark options={scale=0.6}, fill=green!30, fill opacity=0.10, line width=1pt, color=purple]
coordinates {
(90,55.69) (60,42.87) (30,36.48)
(0,56.00) (330,43.18) (300,36.79)
(270,54.09) (240,41.29) (210,34.95)
(180,57.19) (150,44.33) (120,38.08)
(90,55.69)
};
\legend{SLNet-S, SLNet-M, PointNet, PointNet++, DGCNN, PointMLP}
\end{polaraxis}
\end{tikzpicture}
\caption{NetScore and NetScore$^{+}$ across four datasets and two hardware platforms (RTX\,3090 and Jetson Orin Nano), spanning 12 metric–dataset–hardware axes. SLNet-S and SLNet-M consistently occupy the outermost positions, establishing Pareto-optimal deployability across all evaluated configurations.}
\label{fig:radar_plot}
\end{figure}

Existing methods fall into three broad groups, each with a different efficiency bottleneck. Shared-MLP hierarchies, such as PointNet/++~\cite{qi2017pointnet,qi2017pointnet++} and PointMLP~\cite{ma2022rethinking}, are effective but tend to grow in parameter count and latency with model capacity. Graph and kernel based methods, including DGCNN~\cite{wang2019dynamic} and KPConv~\cite{thomas2019kpconv}, rely on repeated neighborhood construction and local aggregation, which can be costly on edge hardware. Transformer based models, such as Point Transformer~\cite{zhao2021point} and Point-BERT~\cite{yu2022point}, often achieve strong accuracy but increase memory use and inference cost because of attention and large learned embeddings~\cite{wu2024point}. At the other extreme, ultra-compact non-parametric models such as NPNet~\cite{saeid2026npnet}, Point-GN~\cite{mohammadi2024point}, and Point-NN~\cite{zhang2023parameter} are efficient but usually trail supervised baselines on harder benchmarks.

\begin{figure*}[t]
  \centering
  \includegraphics[width=0.95\linewidth]{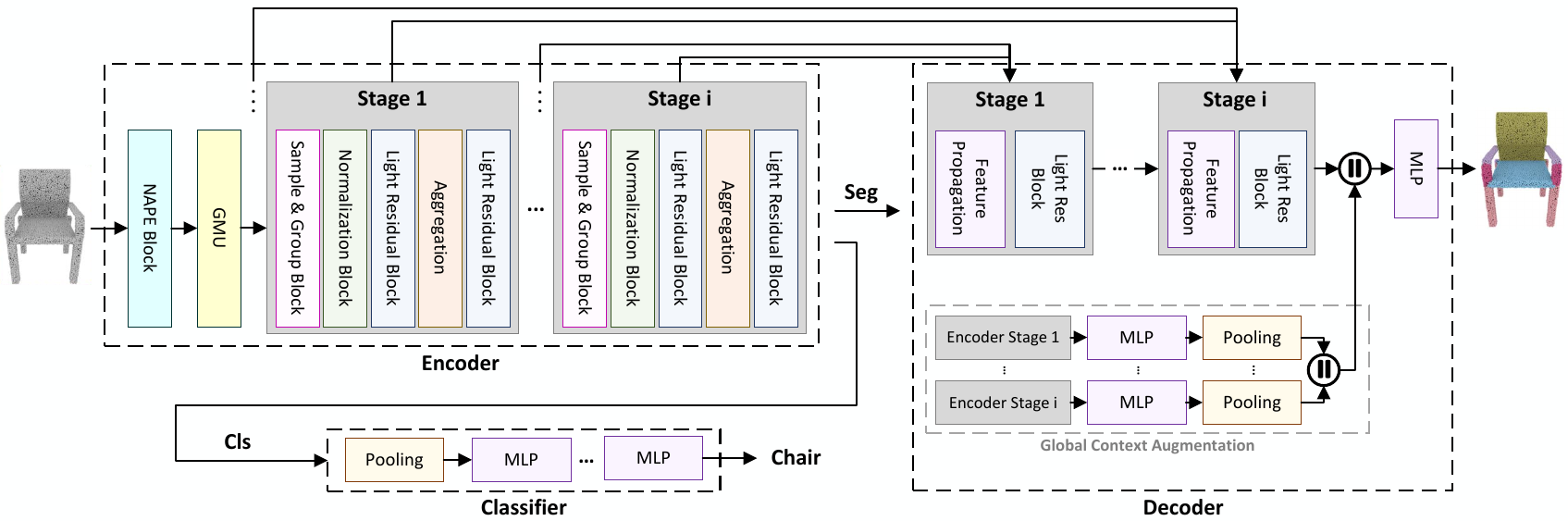}
  \caption{\SLNet architecture: a NAPE+GMU front end, a four-stage hierarchical encoder with FPS, parameter-free normalization, and lightweight residual refinement, and task-specific heads for classification and segmentation.}
  \label{fig:arch}
\end{figure*}

We present \textbf{\SLNet}, a lightweight backbone designed to improve the accuracy-efficiency trade-off in point cloud recognition. It is built around two components: \textbf{NAPE} (Nonparametric Adaptive Point Embedding), which encodes raw XYZ coordinates through an input-adaptive combination of Gaussian radial basis function (RBF) and cosine bases without learned parameters, and \textbf{GMU} (Geometric Modulation Unit), a simple per-channel affine recalibration module with only $2D$ learnable parameters. These components are combined with four hierarchical FPS+kNN stages, parameter-free normalization, and shared residual MLPs. For large-scale indoor scene segmentation, \textbf{\SLNetT} keeps the same overall hierarchy, replaces the NAPE front end with a learned linear projection, and adds local Point Transformer attention at all four encoder stages.

Figure~\ref{fig:radar_plot} summarizes NetScore and NetScore$^{+}$ across 12 metric, dataset, and hardware combinations. In this comparison, SLNet-S shows the strongest overall deployability profile, while SLNet-M also remains consistently competitive across settings. In the few-shot setting, \SLNetM reaches 95.0\% accuracy on 5-way 20-shot classification and 94.5\% on 10-way 20-shot classification without large-scale pretraining.

We also introduce NetScore$^{+}$, a deployment-oriented metric that combines accuracy, parameter count, FLOPs, latency, and memory. Across 12 architectures, it shows strong correlation with measured throughput (Spearman $\rho>0.9$). We evaluate it on ModelNet40, ModelNet-R, ScanObjectNN, ShapeNetPart, and ModelNet40 few-shot, and also report S3DIS results for \SLNet-T.

\noindent\textbf{Contributions:}
\begin{itemize}
  \item We introduce \textbf{NAPE} and \textbf{GMU}, a lightweight combination of nonparametric geometric encoding and ultra-low-cost channel modulation.
  \item We present \textbf{\SLNet} in three variants, \textbf{S}, \textbf{M}, and \textbf{T}, and show that it achieves a strong accuracy-efficiency trade-off across classification, few-shot learning, part segmentation, and scene segmentation.
  \item We introduce \textbf{NetScore$^{+}$}, a deployability metric that incorporates latency and memory in addition to standard efficiency measures.
\end{itemize}
\section{Related Work}
\label{sec:related}

Point cloud backbones span a broad range of design choices, from highly expressive architectures to models developed specifically for efficient deployment.

\noindent\textbf{Shared MLP frameworks.}
PointNet~\cite{qi2017pointnet} introduced the use of shared MLPs and symmetric pooling for point cloud processing, while PointNet++~\cite{qi2017pointnet++} extended this idea with hierarchical grouping to better capture local structure at multiple scales. More recent residual MLP based models, such as PointMLP~\cite{ma2022rethinking}, substantially improve recognition accuracy, but this often comes with increased parameter count and inference cost. Even relatively compact supervised baselines in this family typically exceed 0.7M parameters and 0.9\,GFLOPs at 1k input points~\cite{zhou2024dynamic,sun2024parameter}.

\noindent\textbf{Graph and edge convolution networks.}
Graph based models explicitly model local relationships between points and often provide strong geometric reasoning. DGCNN~\cite{wang2019dynamic}, for example, recomputes $k$NN graphs at each layer and applies edge convolution to capture local interactions. While effective, this repeated neighborhood construction can become expensive as the number of points or network depth increases~\cite{wang2021fast,hou2024graph}, making deployment on edge devices more difficult.

\noindent\textbf{Kernel point convolutions.}
Methods such as PointConv~\cite{wu2019pointconv} and KPConv~\cite{thomas2019kpconv} extend convolution to irregular point sets through continuous or kernel based operators. These approaches are well suited for modeling local geometry, but their kernel parameterization and density compensation mechanisms can add noticeable runtime and memory overhead~\cite{su2020dv,atzmon2018point}.

\noindent\textbf{Transformer and token mixing models.}
Transformer based point cloud models, including PCT~\cite{guo2021pct}, Point Transformer~\cite{zhao2021point}, and masked pretraining approaches such as Point BERT~\cite{yu2022point} and Point MAE~\cite{pang2022masked}, have achieved strong performance on several benchmarks. However, attention operations and large learned embeddings often increase latency and memory consumption~\cite{wu2024point}, which can limit their practicality in resource constrained settings.

\noindent\textbf{Analytic and minimal parameter pipelines.}
A separate line of work focuses on reducing or even removing learnable parameters. Methods such as NPNet~\cite{saeid2026npnet}, Point-NN~\cite{zhang2023parameter}, Point-LN~\cite{salarpour2025pointln}, Point-GN~\cite{mohammadi2024point}, and PointHop++~\cite{zhang2020pointhop++} demonstrate that point cloud recognition can be performed with very small or nearly parameter free pipelines. These methods are highly efficient, but they generally lag behind stronger supervised baselines on more challenging benchmarks~\cite{guo2020deep}.

\section{Methodology}
\label{sec:methodology}

\subsection{Overview}

\SLNet is a four-stage hierarchical backbone for 3D point cloud understanding.
For object-level tasks (\SLNetS/M), the pipeline applies NAPE---a parameter-free geometric encoder---followed by GMU, FPS-based downsampling, parameter-free normalization, and shared residual MLPs.
For scene-level segmentation (\SLNetT), NAPE is replaced with a learned linear projection, and the MLP stages are replaced with local Point Transformer attention.
All three variants share the same hierarchical pipeline based on FPS, kNN grouping, local aggregation, and feature propagation; their key differences are summarized in Table~\ref{tab:variants}, and the overall architecture is shown in Figure~\ref{fig:arch}.

\begin{table}[!bh]
  \centering
  \caption{SLNet variant comparison.}
  \label{tab:variants}
  \setlength{\tabcolsep}{4pt}
  \small
  \begin{tabular}{lcccc}
    \toprule
    Variant & Params & GFLOPs & Front-end & Local op. \\
    \midrule
    \SLNetS & 0.14M & 0.31 & NAPE+GMU & Shared MLP \\
    \SLNetM & 0.55M & 1.22 & NAPE+GMU & Shared MLP \\
    \SLNetT & 2.5M  & 6.5  & Lin.\ proj.\ (6ch) & Point Trans \\
    \bottomrule
  \end{tabular}
\end{table}

\subsection{Nonparametric Adaptive Point Embedding (NAPE)}
\label{sec:nape}

\begin{figure}[!t]
  \centering
  \includegraphics[width=0.99\linewidth]{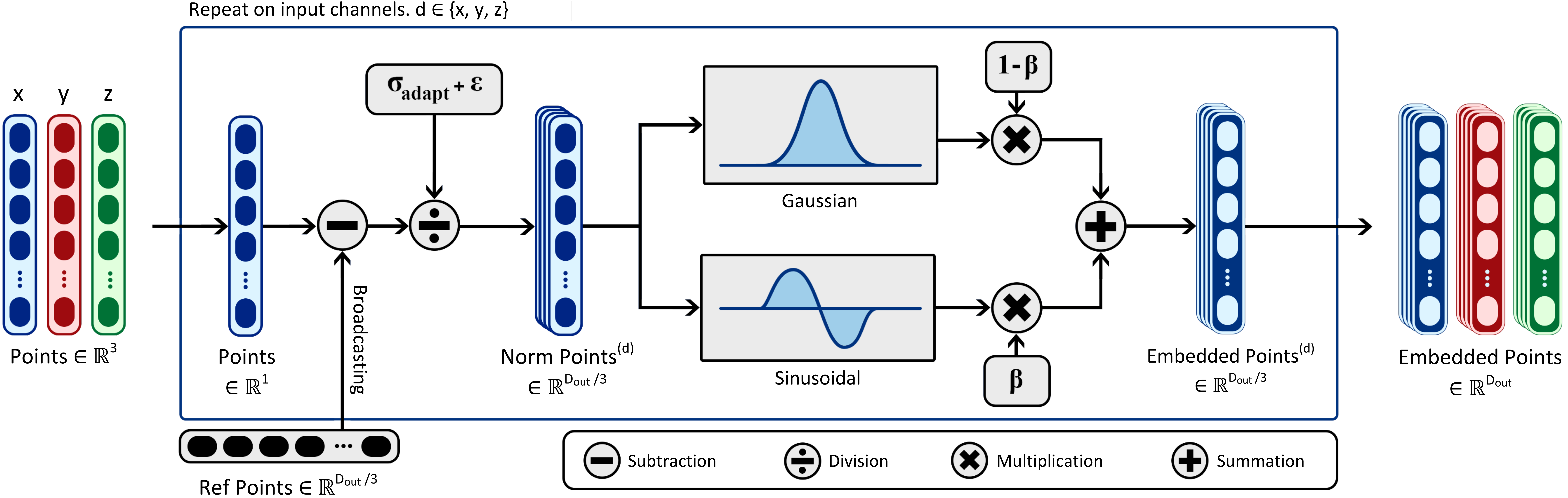}
  \caption{NAPE: raw 3D coordinates mapped via fused Gaussian RBF and cosine bases with data-driven adaptive bandwidth, yielding a parameter-free geometry encoding.}
  \label{fig:nape}
\end{figure}

NAPE is a fully parameter-free block that maps raw XYZ coordinates to a $D$-dimensional feature ($D\!=\!16$ for \SLNetS; $D\!=\!32$ for \SLNetM) through three components.

\noindent\textbf{Global dispersion.}
Object scale is estimated as $\sigma_{\mathrm{global}} = \tfrac{1}{3}(\sigma_x + \sigma_y + \sigma_z)$, the mean per-axis standard deviation across points---a moment-based statistic that is translation-invariant and less sensitive to point-wise noise.

\noindent\textbf{Adaptive bandwidth.}
The kernel width is scaled by object size:
$\sigma_{\mathrm{adapt}} = \sigma_0(1 + \sigma_{\mathrm{global}})$,
with base bandwidth $\sigma_0\!=\!0.4$ fixed across all experiments.

\noindent\textbf{Basis blending.}
For each coordinate axis, Gaussian RBF and cosine bases are evaluated on a uniform grid $\mathbf{g}$ of $\lceil D/3\rceil$ interior points in $(-1, 1)$ (endpoints excluded):
\[
\text{rbf} = \exp\!\Bigl(\!-\tfrac{(x_i - \mathbf{g})^2}{2\,\sigma_{\mathrm{adapt}}^2}\Bigr),\quad
\text{cos} = \cos\!\Bigl(\tfrac{x_i - \mathbf{g}}{\sigma_{\mathrm{adapt}}}\Bigr),
\]
and blended via a sigmoid gate $\beta = \mathrm{sigmoid}\!\left(\gamma(\sigma_{\mathrm{global}} - b)\right)$ ($\gamma\!=\!10$, $b\!=\!0.1$):
\[
\mathbf{e}_i = \beta \cdot \text{rbf} + (1 - \beta) \cdot \text{cos}.
\]
In practice, the Gaussian basis is more localized, while the cosine basis provides smoother responses over larger spatial ranges, with $\beta$ shifting automatically between them based on cloud scale.
Per-axis embeddings are concatenated across the three axes ($3\lceil D/3\rceil$ total features), and a fixed index selection yields the final $D$-dimensional output.
All hyperparameters are fixed across all datasets and model sizes.

\subsection{Geometric Modulation Unit (GMU)}
\label{sec:gmu}

\begin{figure}[t]
  \centering
  \includegraphics[width=0.95\linewidth]{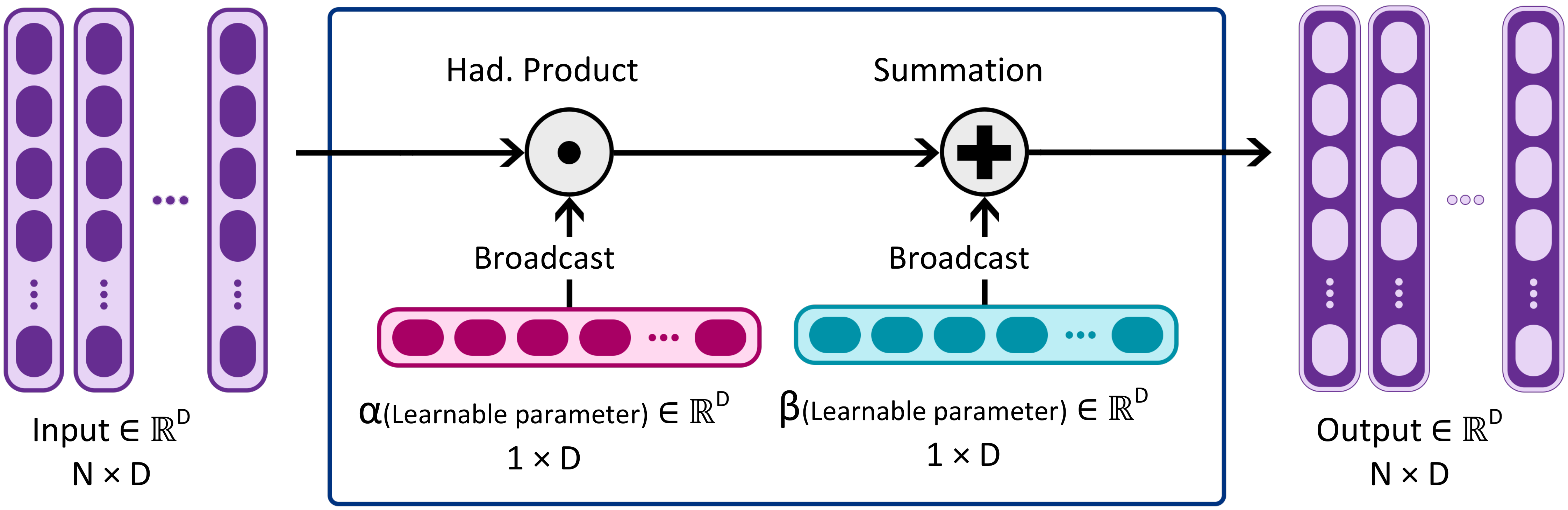}
  \caption{GMU: per-channel affine recalibration of the NAPE output with only $2D$ learnable scalars.}
  \label{fig:arch_gmu}
\end{figure}

GMU applies a per-channel affine transformation after NAPE:
\[
\mathbf{Y}_{b,d,n} = \alpha_d \,\mathbf{X}_{b,d,n} + \beta_d,
\]
where $\alpha_d, \beta_d \in \mathbb{R}$ are learnable scalars ($2D$ parameters total: 32 for \SLNetS, 64 for \SLNetM).
GMU is used where it is empirically beneficial and is omitted for ScanObjectNN based on ablation results.

\subsection{Hierarchical Encoder (\SLNetS/M)}
\label{sec:encoder_sm}

\noindent\textbf{Sampling and grouping.}
Each of four stages uses FPS to select centroids and $k$NN to form local neighborhoods ($K\!=\!32$ for ModelNet40 and ShapeNet; $K\!=\!24$ for ScanObjectNN).
Point counts halve per stage; channel dimensions expand by factors $[2, 2, 2, 1]$.

\noindent\textbf{Parameter-free normalization.}
Relative features are computed without learnable parameters:
\[
\mathbf{H}^1_{ij} = [\mathbf{f}_{ij} \| \mathbf{x}_{ij}] - [\mathbf{f}_i \| \mathbf{x}_i],
\]
providing a relational representation of local geometric context.

\noindent\textbf{Light Residual Block (LRB).}
Features are refined via a shared residual MLP:
\[
\mathbf{Y}_{ij} = \mathrm{ReLU}\!\left(\mathbf{H}^3_{ij} + W_2 \bigl(\mathrm{ReLU}(\mathrm{BN}(W_1 \mathbf{H}^3_{ij}))\bigr)\right),
\]
with channel width ratio $r\!=\!0.25$ (with channel width ratio fixed to $r\!=\!0.25$ based on ablation results).
Stage depths are $[1,1,2,1]$ for ModelNet40 and ShapeNet, and $[1,1,3,1]$ for ScanObjectNN.
Neighbor features are then max-pooled per centroid before the next stage.

\begin{table*}[!t]
\centering
\scriptsize
\caption{\textbf{ModelNet40 classification (supervised).} Metrics: overall accuracy (OA), mean class accuracy (mAcc), parameters (Param, M), FLOPs (G), memory (MB), latency (ms), and composite NetScore / NetScore$^{+}$. Our runs are 3-run averages.}
\label{tab:modelnet40_results}
\begin{adjustbox}{max width=\textwidth}
\begin{tabular}{l|
cc|
cc|
c|c|
cc|
cccc|
c|
cc}
\toprule

\multirow{3}{*}{Method} &
\multicolumn{2}{c|}{\multirow{2}{*}{Paper results}} &
\multicolumn{2}{c|}{\multirow{2}{*}{Our runs}} &
\multicolumn{1}{c|}{\multirow{2}{*}{Param}} &
\multicolumn{1}{c|}{\multirow{2}{*}{FLOPs}} &
\multicolumn{2}{c|}{\multirow{2}{*}{Memory (MB)}} &
\multicolumn{4}{c|}{Inference Time (ms)} &
\multicolumn{1}{c|}{\multirow{3}{*}{NetScore}} &
\multicolumn{2}{c}{\multirow{2}{*}{ NetScore$^{+}$}} \\

&&&&&&&&
& \multicolumn{2}{c|}{3090}
& \multicolumn{2}{c|}{Jetson}
&& \\

& \multicolumn{1}{c}{OA}
& \multicolumn{1}{c|}{mAcc}
& \multicolumn{1}{c}{OA}
& \multicolumn{1}{c|}{mAcc}
& \multicolumn{1}{c|}{(M)}
& \multicolumn{1}{c|}{(G)}
& \multicolumn{1}{c}{3090}
& \multicolumn{1}{c|}{Jetson}
& \multicolumn{1}{c}{1024}
& \multicolumn{1}{c|}{2048}
& \multicolumn{1}{c}{1024}
& \multicolumn{1}{c|}{2048}
& 
& 3090
& Jetson \\

\midrule

PointMLP~\cite{ma2022rethinking} & \textbf{\textcolor{blue}{94.10}} & \textbf{\textcolor{blue}{91.30}} & \underline{\textbf{93.66}} & \underline{\textbf{90.99}} & 13.24 & 15.67 & 86.68 & 105.80 & 4.18 & 4.23 & 64.61 & 65.64 & 55.69 & 42.87 & 36.48 \\
APES (local)~\cite{wu2023attention} & 93.50 & -- & 93.30 & -- & 4.49 & 7.38 & 82.69 & 91.82 & 3.48 & 6.78 & 54.90 & 93.52 & 63.59 & 49.84 & 43.92 \\
APES (global)~\cite{wu2023attention} & 93.80 & -- & 93.23 & -- & 4.49 & 5.49 & 82.69 & 91.82 & 2.50 & 4.89 & 37.28 & 65.97 & 64.86 & 51.83 & 45.95 \\
PointNet++ (msg)~\cite{qi2017pointnet++} & -- & -- & 92.51 & 89.87 & 1.75 & 4.00 & 99.87 & 108.99 & 4.56 & 5.92 & 209.96 & 218.54 & 70.21 & 56.35 & 48.32 \\
DGCNN~\cite{wang2019dynamic} & 92.90 & 90.20 & 92.82 & 89.09 & 1.81 & 2.69 & 78.09 & 87.22 & 1.56 & 3.90 & 25.71 & 59.87 & 71.84 & 59.42 & 53.25 \\
PointNet++ (ssg)~\cite{qi2017pointnet++} & 90.70 & -- & 92.31 & 89.65 & 1.48 & 0.86 & 25.93 & 35.15 & 2.26 & 2.73 & 171.04 & 181.25 & 77.59 & 68.34 & 58.57 \\
CurveNet~\cite{xiang2021walk} & 93.80 & -- & 93.38 & -- & 2.14 & \underline{\textbf{0.33}} & 21.33 & 30.45 & 3.04 & 8.17 & 108.97 & 403.57 & 80.34 & 69.14 & 59.89 \\
PointNet~\cite{qi2017pointnet} & 89.20 & 86.20 & 90.04 & 86.52 & 3.47 & 0.45 & 22.08 & 31.20 & \textbf{\textcolor{blue}{0.19}} & \textbf{\textcolor{blue}{0.35}} & \textbf{\textcolor{blue}{3.85}} & \textbf{\textcolor{blue}{5.40}} & 76.27 & 71.86 & 65.13 \\
PointMLP (elite)~\cite{ma2022rethinking} & 93.60 & 90.90 & 93.28 & \underline{\textbf{90.99}} & 0.72 & 0.91 & \underline{\textbf{18.56}} & \underline{\textbf{28.56}} & 1.18 & 1.22 & 25.28 & 25.99 & 80.64 & \underline{\textbf{73.87}} & \underline{\textbf{66.29}} \\
\midrule
\rowcolor{lightgreen} SLNet-S & 93.64 & 89.46 & 93.64 & 89.46 & \textbf{\textcolor{blue}{0.14}} & \textbf{\textcolor{blue}{0.31}} & \textbf{\textcolor{blue}{11.49}} & \textbf{\textcolor{blue}{21.30}} & \underline{\textbf{0.74}} & \underline{\textbf{0.76}} & \underline{\textbf{17.31}} & \underline{\textbf{18.04}} & \textbf{\textcolor{blue}{92.42}} & \textbf{\textcolor{blue}{87.71}} & \textbf{\textcolor{blue}{79.50}} \\
\rowcolor{lightgreen} SLNet-M & \underline{\textbf{93.92}} & \underline{\textbf{91.10}} & \textbf{\textcolor{blue}{93.92}} & \textbf{\textcolor{blue}{91.10}} & \underline{\textbf{0.55}} & 1.22 & 23.97 & 33.00 & 1.39 & 1.41 & 29.48 & 30.26 & \underline{\textbf{80.66}} & 73.01 & 65.67 \\

\bottomrule
\end{tabular}
\end{adjustbox}
\end{table*}

\begin{table}[t]
\centering
\scriptsize
\caption{\textbf{ModelNet\mbox{-}R classification (supervised).} Param, FLOPs, Mem, and Time match Table~\ref{tab:modelnet40_results} and are omitted here.}
\label{tab:modelnetR_results}
\begin{adjustbox}{max width=\columnwidth}
\begin{tabular}{l | c | c | c | c | c } 
\toprule
\multirow{2}{*}{Method} & \multicolumn{1}{c}{\multirow{2}{*}{OA}}
& \multicolumn{1}{c|}{\multirow{2}{*}{mOAcc}}
& \multicolumn{1}{c|}{\multirow{2}{*}{NetScore}}
& \multicolumn{2}{c}{NetScore$^{+}$} \\
 &  &  &  & 3090 & Jetson \\
\midrule
PointMLP~\cite{ma2022rethinking} & \textbf{\textcolor{blue}{95.33}} & \textbf{\textcolor{blue}{94.30}} & 56.00 & 43.18 & 36.79 \\
PointNet++ (msg)~\cite{qi2017pointnet++} & 94.06 & 91.80 & 70.50 & 56.64 & 48.61 \\
DGCNN~\cite{wang2019dynamic} & 94.03 & 92.64 & 72.06 & 59.65 & 53.47 \\
PointNet++ (ssg)~\cite{qi2017pointnet++} & 94.02 & 92.40 & 77.91 & 68.66 & 58.89 \\
CurveNet~\cite{xiang2021walk} & 94.12 & 92.65 & 80.48 & 69.27 & 60.03 \\
PointNet~\cite{qi2017pointnet} & 91.39 & 88.79 & 76.53 & 72.12 & 65.39 \\
\midrule
\rowcolor{lightgreen} SLNet-S & \underline{\textbf{94.53}} & 92.21 & \textbf{\textcolor{blue}{92.59}} & \textbf{\textcolor{blue}{87.87}} & \textbf{\textcolor{blue}{79.66}} \\
\rowcolor{lightgreen} SLNet-M & 94.81 & \underline{\textbf{93.76}} & \underline{\textbf{80.83}} & \underline{\textbf{73.18}} & \underline{\textbf{65.83}} \\
\bottomrule
\end{tabular}
\end{adjustbox}
\end{table}

\subsection{Task-Specific Heads}

\noindent\textbf{Classification (\SLNetS/M).}
Global adaptive max-pooling yields a fixed-length descriptor, fed to a two-layer MLP classifier.

\noindent\textbf{Part segmentation (\SLNetS/M).}
A U-Net decoder restores resolution via four FP blocks with inverse-distance-weighted ($k\!=\!3$) interpolation and skip connections.
Multi-scale GMP fusion concatenates pooled features from all five encoder levels with a class-label embedding and the final decoder output before per-point classification.

\noindent\textbf{Semantic segmentation (\SLNetT).}
\SLNetT replaces NAPE with a learned linear projection ($6 \to 64$ channels, LayerNorm, ReLU) to accommodate XYZ+RGB inputs, followed by four encoder stages with channel dimensions $[64, 128, 256, 512]$ and local Point Transformer v1 attention~\cite{zhao2021point}: subtraction-based weights $\mathbf{w} = \mathrm{MLP}(\mathbf{q} - \mathbf{k} + \mathbf{pe})$ (softmax over $k\!=\![16,16,32,64]$ neighbors) with a shared relative position encoding branch.
GroupNorm enables stable training at batch size~1 on dense scene inputs.
The decoder is a lightweight FP stack (dims $[256, 128, 64, 64]$, no GMP fusion), trained with weighted cross-entropy (inverse-square-root class weights, label smoothing $\epsilon\!=\!0.1$) to address S3DIS class imbalance.
\section{Experiments}
\label{sec:experiments}

We evaluate \SLNet on six benchmarks spanning object classification, part segmentation, few-shot learning, and large-scale semantic segmentation.

\subsection{Setup}
\label{sec:setup}

\noindent\textbf{Object tasks (\SLNetS/M).}
Experiments were conducted on consumer GPUs (NVIDIA GTX 1080–RTX 3090) under Ubuntu 22.04 with CUDA 11.8. Inputs: 1024 points for classification/few-shot, 2048 points with normals for ShapeNetPart. ModelNet40/R and ScanObjectNN: 300 epochs, SGD (lr 0.1/0.01), cosine decay, EMA. ShapeNetPart: 350 epochs, AdamW (lr 0.003). FLOPs were computed with FVCore.

Metrics include trainable parameters, GFLOPs, peak GPU memory, and average inference time per sample. Measurements follow the supplied routines for counting parameters, computing GFLOPs, measuring peak memory, and timing inference with warmup. Input sizes match experimental settings (Table~\ref{tab:eff_hw}).

\begin{table}[!t]
  \centering
  \caption{Hardware and software used for efficiency benchmarking.}
  \label{tab:eff_hw}
  \footnotesize
  \setlength{\tabcolsep}{3pt}
  \begin{tabularx}{\linewidth}{lccXc}
    \toprule
    Device & Batch & Workers & Software stack & Memory \\
    \midrule
    RTX 3090 & 32 & 6 & Ubuntu 22.04, CUDA 11.8 & 25.3\,GB \\
    Jetson Orin Nano & 4 & 4 & JetPack 6.2.2, CUDA 12.6, Python 3.10 & 8\,GB \\
    \bottomrule
  \end{tabularx}
\end{table}


\noindent\textbf{Scene segmentation (\SLNetT).}
Trained on an RTX 5090 with AdamW ($\eta\!=\!9\!\times\!10^{-4}$, $\lambda\!=\!2\!\times\!10^{-3}$), cosine schedule with 15-epoch warm-up, mixed precision, and EMA ($\rho\!=\!0.999$).
Scene crops: $N\!=\!16{,}384$ points, radius 2.5\,m, 75 crops/room; 6D input (XYZ+RGB).
Validation every 5 epochs via sliding-window inference (stride 0.75\,m).

\noindent\textbf{Edge profiling.} All variants were benchmarked on a Jetson Orin Nano with 1024 input points and batch size 4; the hardware and software settings are summarized in Table~\ref{tab:eff_hw}.


\subsection{Metrics}
\label{sec:metrics}

Classification: overall accuracy (OA), mean class accuracy (mAcc).
Part segmentation: instance- and class-average IoU (ins-IoU, cls-IoU).
Scene segmentation: mIoU, mAcc, OA.
Efficiency: parameters $p$\,(M), FLOPs $m$\,(G), peak memory $r$\,(MB), latency $t$\,(ms).

\begin{table*}[t]
\centering
\scriptsize
\caption{\textbf{ScanObjectNN classification (supervised).} Metrics: OA, mAcc, Param (M), FLOPs (G), Mem (MB), Time (ms), and NetScore/NetScore$^{+}$.}
\label{tab:scanobjectnn_results}
\begin{adjustbox}{max width=\textwidth}
\begin{tabular}{l | cc | c|c | cc | cccc | c | cc}
\toprule

\multirow{3}{*}{Method} &
\multicolumn{2}{c|}{\multirow{2}{*}{Paper results}} &
\multicolumn{1}{c|}{\multirow{2}{*}{Param}} &
\multicolumn{1}{c|}{\multirow{2}{*}{FLOPs}} &
\multicolumn{2}{c|}{\multirow{2}{*}{Memory (MB)}} &
\multicolumn{4}{c|}{Inference Time (ms)} &
\multicolumn{1}{c|}{\multirow{3}{*}{NetScore}} &
\multicolumn{2}{c}{\multirow{2}{*}{ NetScore$^{+}$}} \\

&&&&&&
& \multicolumn{2}{c|}{3090}
& \multicolumn{2}{c|}{Jetson}
&& \\

& \multicolumn{1}{c}{OA}
& \multicolumn{1}{c|}{mAcc}
& \multicolumn{1}{c|}{(M)}
& \multicolumn{1}{c|}{(G)}
& \multicolumn{1}{c}{3090}
& \multicolumn{1}{c|}{Jetson}
& \multicolumn{1}{c}{1024}
& \multicolumn{1}{c|}{2048}
& \multicolumn{1}{c}{1024}
& \multicolumn{1}{c|}{2048}
&
& 3090
& Jetson \\

\midrule

PointNet++ (msg)~\cite{qi2017pointnet++} & -- & -- & 1.74 & 4.00 & 99.84 & 108.97 & 4.49 & 5.83 & 211.28 & 213.84 & -- & -- & -- \\
CurveNet~\cite{xiang2021walk} & -- & -- & 2.13 & \underline{\textbf{0.33}} & 21.28 & 30.40 & 3.02 & 8.06 & 115.16 & 405.10 & -- & -- & -- \\
APES (global)~\cite{wu2023attention} & -- & -- & 4.49 & 5.49 & 82.67 & 91.79 & 2.50 & 4.88 & 37.31 & 66.15 & -- & -- & -- \\
APES (local)~\cite{wu2023attention} & -- & -- & 4.49 & 7.38 & 82.67 & 91.79 & 3.48 & 6.79 & 55.11 & 93.65 & -- & -- & -- \\
PointMLP~\cite{ma2022rethinking} & \textbf{\textcolor{blue}{85.40}} & \textbf{\textcolor{blue}{83.90}} & 13.24 & 15.67 & 86.68 & 99.80 & 4.12 & 4.19 & 64.98 & 67.51 & 54.09 & 41.29 & 34.95 \\
DGCNN~\cite{wang2019dynamic} & 78.10 & 73.60 & 1.80 & 2.69 & 78.07 & 87.20 & 1.55 & 3.90 & 25.72 & 59.87 & 68.85 & 56.44 & 50.27 \\
PointNet++ (ssg)~\cite{qi2017pointnet++} & 77.90 & 75.40 & 1.47 & 0.86 & 25.91 & 35.03 & 2.23 & 2.73 & 175.69 & 180.70 & 74.66 & 65.42 & 55.66 \\
PointNet~\cite{qi2017pointnet} & 68.20 & 63.40 & 3.47 & 0.45 & 22.05 & 31.18 & \textbf{\textcolor{blue}{0.19}} & \textbf{\textcolor{blue}{0.35}} & \textbf{\textcolor{blue}{4.66}} & \textbf{\textcolor{blue}{5.31}} & 71.45 & 67.04 & 60.35 \\
PointMLP (elite)~\cite{ma2022rethinking} & 83.80 & 81.80 & 0.72 & 0.91 & 18.56 & 27.68 & 1.15 & 1.20 & 25.37 & 26.23 & 78.78 & 72.04 & 64.48 \\
\midrule
\rowcolor{lightgreen} SLNet-S & 83.45 & 81.45 & \textbf{\textcolor{blue}{0.12}} & \textbf{\textcolor{blue}{0.26}} & \textbf{\textcolor{blue}{8.75}} & \textbf{\textcolor{blue}{17.87}} & \underline{\textbf{0.64}} & \underline{\textbf{0.66}} & \underline{\textbf{16.36}} & \underline{\textbf{16.79}} & \textbf{\textcolor{blue}{91.76}} & \textbf{\textcolor{blue}{87.96}} & \textbf{\textcolor{blue}{79.38}} \\
\rowcolor{lightgreen} SLNet-M & \underline{\textbf{84.25}} & \underline{\textbf{82.86}} & \underline{\textbf{0.48}} & 1.02 & \underline{\textbf{18.28}} & \underline{\textbf{27.37}} & 1.22 & 1.24 & 27.30 & 27.57 & \underline{\textbf{80.12}} & \underline{\textbf{73.34}} & \underline{\textbf{65.73}} \\

\bottomrule
\end{tabular}
\end{adjustbox}
\end{table*}

\begin{table*}[t]
\centering
\scriptsize
\caption{\textbf{ShapeNetPart part segmentation (supervised).} Metrics: instance IoU (ins-IoU), class-average IoU (cls-IoU), Param (M), FLOPs (G), Mem (MB), Time (ms), and NetScore/NetScore$^{+}$.}
\label{tab:shapenet_results}
\begin{adjustbox}{max width=\textwidth}
\begin{tabular}{l | cc | c | cc | cccc | cccc | c | cc}
\toprule

\multirow{3}{*}{Method} &
\multicolumn{2}{c|}{\multirow{2}{*}{Paper results}} &
\multicolumn{1}{c|}{\multirow{2}{*}{Param}} &
\multicolumn{2}{c|}{\multirow{2}{*}{FLOPs (G)}} &
\multicolumn{4}{c|}{Memory (MB)} &
\multicolumn{4}{c|}{Inference Time (ms)} &
\multicolumn{1}{c|}{\multirow{3}{*}{NetScore}} &
\multicolumn{2}{c}{\multirow{2}{*}{ NetScore$^{+}$}} \\

&&&&&
& \multicolumn{2}{c|}{3090}
& \multicolumn{2}{c|}{Jetson}
& \multicolumn{2}{c|}{3090}
& \multicolumn{2}{c|}{Jetson}
&& \\

& \multicolumn{1}{c}{ins-IoU}
& \multicolumn{1}{c|}{cls-IoU}
& \multicolumn{1}{c|}{(M)}
& \multicolumn{1}{c}{1024}
& \multicolumn{1}{c|}{2048}
& \multicolumn{1}{c}{1024}
& \multicolumn{1}{c|}{2048}
& \multicolumn{1}{c}{1024}
& \multicolumn{1}{c|}{2048}
& \multicolumn{1}{c}{1024}
& \multicolumn{1}{c|}{2048}
& \multicolumn{1}{c}{1024}
& \multicolumn{1}{c|}{2048}
&
& 3090
& Jetson \\

\midrule

PointNet++ (msg)~\cite{qi2017pointnet++} & -- & -- & 1.74 & 4.76 & 4.82 & 99.83 & 103.84 & 107.95 & 111.96 & 5.04 & 6.64 & 211.25 & 210.46 & -- & -- & -- \\
APES (global)~\cite{wu2023attention} & 85.80 & 83.70 & 1.98 & -- & 15.57 & -- & 140.38 & -- & 148.51 & -- & 7.47 & -- & 101.55 & 62.45 & 47.35 & 41.56 \\
APES (local)~\cite{wu2023attention} & 85.60 & 83.10 & 1.98 & -- & 18.37 & -- & 140.38 & -- & 148.51 & -- & 9.38 & -- & 128.43 & 61.69 & 46.09 & 40.29 \\
CurveNet~\cite{xiang2021walk} & \textbf{\textcolor{blue}{86.60}} & -- & 5.53 & -- & 2.56 & -- & 97.90 & -- & 106.03 & -- & 10.32 & -- & 280.34 & 65.98 & 50.96 & 43.62 \\
DGCNN~\cite{wang2019dynamic} & 85.20 & 82.30 & 1.46 & 2.34 & 4.96 & 77.38 & 149.14 & 86.50 & 158.27 & \underline{\textbf{1.94}} & 5.29 & \underline{\textbf{29.56}} & 64.81 & 68.62 & 54.13 & 48.56 \\
PointMLP~\cite{ma2022rethinking} & \underline{\textbf{86.10}} & \textbf{\textcolor{blue}{84.60}} & 16.76 & 6.05 & 6.26 & 113.46 & 113.73 & 121.65 & 121.23 & 3.15 & 3.28 & 57.61 & 54.76 & 57.19 & 44.33 & 38.08 \\
PointNet~\cite{qi2017pointnet} & 83.70 & 80.40 & 8.34 & 2.90 & 5.79 & 97.26 & 157.72 & 83.93 & 125.39 & \textbf{\textcolor{blue}{0.49}} & \textbf{\textcolor{blue}{1.02}} & \textbf{\textcolor{blue}{8.60}} & \textbf{\textcolor{blue}{11.54}} & 60.07 & 49.03 & 44.27 \\
PointNet++ (ssg)~\cite{qi2017pointnet++} & 85.10 & 81.90 & \underline{\textbf{1.41}} & \underline{\textbf{1.05}} & \underline{\textbf{1.13}} & \underline{\textbf{26.43}} & 58.19 & \underline{\textbf{34.55}} & \textbf{\textcolor{blue}{42.33}} & 2.67 & 3.35 & 174.82 & 190.38 & \underline{\textbf{75.19}} & \underline{\textbf{63.74}} & \underline{\textbf{55.66}} \\
\midrule
\rowcolor{lightgreen} SLNet-S & 85.21 & 83.99 & \textbf{\textcolor{blue}{1.24}} & \textbf{\textcolor{blue}{0.83}} & \textbf{\textcolor{blue}{0.91}} & \textbf{\textcolor{blue}{21.99}} & \textbf{\textcolor{blue}{38.12}} & \textbf{\textcolor{blue}{30.09}} & \underline{\textbf{46.24}} & 2.17 & \underline{\textbf{2.49}} & 38.69 & \underline{\textbf{40.88}} & \textbf{\textcolor{blue}{76.70}} & \textbf{\textcolor{blue}{66.81}} & \textbf{\textcolor{blue}{60.31}} \\
\rowcolor{lightgreen} SLNet-M & 85.53 & \underline{\textbf{84.45}} & 1.90 & 2.25 & 2.33 & 40.15 & \underline{\textbf{41.33}} & 48.27 & 49.46 & 3.39 & 3.72 & 58.59 & 60.66 & 70.81 & 59.88 & 53.43 \\

\bottomrule
\end{tabular}
\end{adjustbox}
\end{table*}


\noindent\textbf{NetScore and NetScore$^{+}$.}
To quantify accuracy–efficiency trade-offs as a single deployability metric, we adopt NetScore~\cite{wong2018netscore} and define:

{\scriptsize
\begin{equation}\label{eq:netscore}
\mathrm{NetScore},\,\mathrm{NetScore}^{+} = 20\log_{10} 
\frac{a^2}{\sqrt{p m}\,\sqrt[4]{t r}^{\,\delta}},\;\;
\delta = \begin{cases}0 & \text{for NetScore},\\1 & \text{for NetScore$^+$}\end{cases}
\end{equation}
}

where $a$ is accuracy in percent (not fractional), $p$ is number of parameters, $m$ FLOPs, $t$ inference time, and $r$ memory footprint. Higher is better. Only methods with reproducible hardware metrics are reported.

\subsection{Classification}

\noindent\textbf{ModelNet40.}
As shown in Table~\ref{tab:modelnet40_results}, \SLNetS (0.14M, 0.31\,G) achieves 93.64\% OA, surpassing PointMLP-elite (93.28\%) with $5\times$ fewer parameters and attaining the highest NetScore (92.42) and NetScore$^{+}$ (87.71) among all evaluated methods.
\SLNetM (0.55M, 1.22\,G) reaches 93.92\%, exceeding PointMLP in our runs (93.66\%) at $24\times$ fewer parameters and $13\times$ fewer FLOPs, and matching it in mAcc (91.1\%).

\noindent\textbf{ModelNet‑R.}
ModelNet‑R is a refined version of the widely used ModelNet40 classification benchmark, with corrected labels, removal of near‑2D and ambiguous samples, and improved class distinctions to yield a more reliable evaluation dataset~\cite{saeid2025enhancing}. Table~\ref{tab:modelnetR_results} shows consistent gains on this re‑annotated benchmark: \SLNetS reaches 94.53\% and \SLNetM 94.81\%, with NetScore and NetScore$^{+}$ rankings preserved.

\noindent\textbf{ScanObjectNN (PB-T50-RS).}
This split includes background clutter, partial occlusion, random rotations, and point sparsity. \SLNetM achieves 84.25\% OA with $28\times$ fewer parameters and $15\times$ fewer FLOPs than PointMLP (85.40\%), while remaining within 1.15 percentage points in accuracy. \SLNetS reaches 83.45\% OA and achieves stronger NetScore and NetScore$^{+}$ than PointMLP-elite (Table~\ref{tab:scanobjectnn_results}).

\subsection{Part Segmentation on ShapeNetPart}
\SLNetS reaches 85.21\% ins-IoU, and posts the highest NetScore$^{+}$ (66.81 / 60.31) of all evaluated methods (Table~\ref{tab:shapenet_results}).
\SLNetM attains 85.53\% ins-IoU at only 3.72\,ms per cloud (2048 points), $2.8\times$ faster than CurveNet at just 1.1\,pp lower IoU.
Both variants demonstrate that the U-Net decoder with multi-scale GMP fusion preserves fine part boundaries while sustaining single-digit millisecond inference.

\subsection{Semantic Segmentation on S3DIS}

Table~\ref{tab:S3DIS_results} shows that \SLNetT achieves 58.2\% mIoU, 85.3\% OA, and 65.4\% mAcc on S3DIS Area~5 with only 2.5M parameters and 6.5\,GFLOPs. Although its absolute mIoU is lower than heavier transformer baselines such as PT (70.4\%) and ConDAF (73.5\%), \SLNetT attains a higher NetScore (58.5 vs.\ 57.5 for PT and 54.8 for FPT), indicating a stronger accuracy-efficiency trade-off under a much smaller model budget. This result reflects the design goal of \SLNetT: prioritizing deployment efficiency under a $\leq$2.5M parameter budget, which is approximately $3\times$ smaller than PT, $5\times$ smaller than ConDAF, and $17\times$ smaller than PT~V3. Within this budget, replacing MLP stages with local attention improves mIoU by 14.4 percentage points over the MLP-only baseline (§\ref{sec:ablation}), highlighting the importance of local relational modeling for scene-level understanding.

\subsection{Few-Shot Classification}

Table~\ref{tab:fewshot_results} reports performance on ModelNet40 under 5-way and 10-way episodic protocols.
\SLNetM achieves 95.0\% (5-way 20-shot) and \textbf{94.0\%} (10-way 20-shot); \SLNetS reaches \textbf{93.5\%} in the same 10-way 20-shot setting.
Notably, in the 10-way 20-shot configuration both variants—despite being parametric—surpass all non-parametric methods, including NPNet~\cite{saeid2026npnet} (87.6\%), improving by about 6 percentage points (SLNet-M: +6.4\,pp; SLNet-S: +5.9\,pp).

\begin{table}[t]
\centering
\scriptsize

\caption{\textbf{S3DIS Area~5 semantic segmentation}. SLNet-T (2.5M params) vs. transformer-based models. Despite lower mIoU, it achieves the highest NetScore (58.5), indicating the best accuracy per parameter.}
\label{tab:S3DIS_results}

\begin{adjustbox}{max width=\columnwidth}
\begin{tabular}{l | c | c | c | c | c | c}
\toprule
\textbf{Method} & \textbf{OA} & \textbf{mAcc} & \textbf{mIoU} & \textbf{Param (M)} & \textbf{FLOPs (G)} & \textbf{NetScore} \\
\midrule

PT & 90.8 & 76.5 & 70.4 & \underline{\textbf{7.8}} & \textbf{\textcolor{blue}{5.6}} & \underline{\textbf{57.5}} \\
ST & \underline{\textbf{91.5}} & 78.1 & 72.0 & -- & -- & -- \\
PT V2 & 91.1 & 77.9 & 71.6 & 12.8 & -- & -- \\
ConDAF & \textbf{\textcolor{blue}{91.6}} & \textbf{\textcolor{blue}{78.9}} & \textbf{\textcolor{blue}{73.5}} & 12.8 & -- & -- \\
PT V3 & -- & -- & \underline{\textbf{73.1}} & 42.6 & -- & -- \\ 
FPT & \underline{\textbf{91.5}} & \underline{\textbf{78.5}} & 72.2 & 10.9 & 8.3 & 54.8 \\

\midrule
\rowcolor{lightgreen} SLNet-T & 85.3 & 65.4 & 58.2 & \textbf{\textcolor{blue}{2.5}} & \underline{\textbf{6.5}} & \textbf{\textcolor{blue}{58.5}} \\

\bottomrule
\end{tabular}
\end{adjustbox}

\begin{tablenotes}\scriptsize
\item Results of prior methods are primarily taken from~\cite{he2025pointdiffuse} and supplemented with values reported in the respective original papers when necessary.
\end{tablenotes}

\end{table}

\begin{table}[t]
\centering
\scriptsize
\caption{\textbf{Few-shot classification (\%) on ModelNet40.} 5-way/10-way, 10-shot and 20-shot. Baselines from Sharma et al.~\cite{sharma2020self}.}
\label{tab:fewshot_results}
\begin{adjustbox}{max width=\columnwidth}
\begin{tabular}{l | c|c | c|c}
\toprule
\multirow{2}{*}{Method} & \multicolumn{2}{c}{5-way} & \multicolumn{2}{c}{10-way} \\
 & 10-shot & 20-shot & 10-shot & 20-shot \\
\midrule
\multicolumn{5}{l}{Parametric Methods} \\
FoldingNet~\cite{yang2018foldingnet} & 33.4 & 35.8 & 18.6 & 15.4 \\
DGCNN~\cite{wang2019dynamic} & 31.6 & 40.8 & 19.9 & 16.9 \\
PointNet~\cite{qi2017pointnet} & 52.0 & 57.8 & 46.6 & 35.2 \\
PointNet++~\cite{qi2017pointnet++} & 38.5 & 42.4 & 23.0 & 18.8 \\
3D-GAN~\cite{wu2016learning} & 55.8 & 65.8 & 40.3 & 48.4 \\
PointCNN~\cite{li2018pointcnn} & 65.4 & 68.6 & 46.6 & 50.0 \\
\midrule
\multicolumn{5}{l}{Non-Parametric Methods} \\
Point-NN~\cite{zhang2023parameter} & 88.8 & 90.9 & 79.9 & 84.9 \\
Point-GN~\cite{salarpour2024pointgn} & \underline{\textbf{90.7}} & 90.9 & \underline{\textbf{81.6}} & 86.4 \\
NPNet~\cite{saeid2026npnet} & \textbf{\textcolor{blue}{92.0}} & \underline{\textbf{93.2}} &\textbf{\textcolor{blue}{82.5}} & 87.6 \\
\midrule
\rowcolor{lightgreen} \multicolumn{5}{l}{Our Method (Parametric)} \\
\rowcolor{lightgreen} SLNet-S & 84.0 & 89.0 & 75.5 & \underline{\textbf{93.5}} \\
\rowcolor{lightgreen} SLNet-M & 89.0 & \textbf{\textcolor{blue}{95.0}} & 80.0 & \textbf{\textcolor{blue}{94.0}} \\
\bottomrule
\end{tabular}
\end{adjustbox}
\end{table}

\subsection{Ablation Study}
\label{sec:ablation}

\noindent\textbf{Embedding and GMU placement (\SLNetS/M).}
Table~\ref{tab:ablation_embd} compares embedding choices and GMU insertion points on ModelNet40.
NAPE alone (without GMU) outperforms learned MLP, pure Gaussian, and pure cosine embeddings by about 0.9--1.1 percentage points, suggesting that adaptive basis blending provides a more effective geometric encoding than any single basis alone.
Adding GMU with scale-then-shift ($\alpha\beta$) immediately after the embedding recovers a further 0.5\,pp for both \SLNetS (93.13\,$\to$\,93.64\%) and \SLNetM (93.44\,$\to$\,93.92\%), while placing GMU after sampling\,\&\,grouping or using the alternative $\beta\alpha$ order consistently underperforms; double GMU also fails to improve over single post-embedding placement.

\noindent\textbf{Neighborhood size.}
Table~\ref{tab:ablation_knn} sweeps $K \in \{16,24,32,64\}$.
$K\!=\!32$ is optimal for ModelNet40 (clean, uniform sampling), while $K\!=\!24$ is optimal for ScanObjectNN (real-world clutter), consistent with the intuition that over-large neighborhoods on noisy data introduce irrelevant context.
Performance degrades noticeably at $K\!=\!64$ on both datasets ($-$0.75\,pp on \SLNetM/ModelNet40), so the default was tuned per dataset accordingly.

\noindent\textbf{LRB channel width.}
Table~\ref{tab:ablation_residual} evaluates the residual bottleneck ratio $r \in \{\div8, \div4, \div2, 1\}$.
$r\!=\!\div4$ is the clear optimum on both datasets: $\div8$ loses 1.6\,pp on \SLNetS while saving only 0.03M parameters, and expanding to $\div2$ or $r\!=\!1$ adds parameters with no accuracy gain.
This suggests that the $\div4$ bottleneck provides a useful regularization effect without becoming a limiting capacity constraint.

\begin{table}[t!]
\centering
\scriptsize

\caption{Effect of embedding choice and GMU placement on ModelNet40 OA\,(\%). NAPE with post-embedding $\alpha\beta$ modulation is optimal across both model sizes.}

\label{tab:ablation_embd}

\begin{adjustbox}{max width=\columnwidth}

\begin{tabular}{ccccc}  
\toprule
\makecell{Embedding} & 
\makecell{GMU after\\Embedding} & 
\makecell{GMU after\\Sampling\&Grouping} & 
\makecell{SLNet‑\\S} & 
\makecell{SLNet‑\\M} \\
\midrule
NAPE      & —                    & —                             & 93.13    & 93.44    \\
NAPE      & $\alpha\beta$        & —                             & \textbf{93.64}    & \textbf{93.92}    \\
NAPE      & —                    & $\alpha\beta$                 & 93.27    & 93.52    \\
NAPE      & $\beta\alpha$        & —                             & 93.19    & 93.47    \\
NAPE      & —                    & $\beta\alpha$                 & 92.63    & 92.88    \\
NAPE      & $\alpha\beta$        & $\alpha\beta$                 & 93.40    & 93.68     \\
NAPE      & $\beta\alpha$        & $\beta\alpha$                 & 93.23    & 93.51    \\
\midrule
mlp       & $\alpha\beta$        & —                             & 92.71    & 92.93    \\
mlp       & —                    & $\alpha\beta$                 & 92.49    & 92.59    \\
mlp       & $\alpha\beta$        & $\alpha\beta$                 & 92.60    & 92.78    \\
\midrule
gaussian  & $\alpha\beta$        & —                             & 92.57    & 92.82    \\
gaussian  & —                    & $\alpha\beta$                 & 92.31    & 92.74    \\
gaussian  & $\alpha\beta$        & $\alpha\beta$                 & 92.40    & 92.69    \\
\midrule
cosine    & $\alpha\beta$        & —                             & 92.55    & 92.77    \\
cosine    & —                    & $\alpha\beta$                 & 92.34    & 92.65    \\
cosine    & $\alpha\beta$        & $\alpha\beta$                 & 92.65    & 92.81    \\

\bottomrule

\end{tabular}
\end{adjustbox}
\end{table}

\begin{table}[b!]
\centering
\scriptsize

\caption{Sensitivity to neighbourhood size $K$ on ModelNet40 and ScanObjectNN. Optimal $K$ differs across datasets due to point density and clutter.}
\label{tab:ablation_knn}

\begin{adjustbox}{max width=\columnwidth}
\begin{tabular}{c|c|cccccc}

\toprule
\textbf{Dataset} & \textbf{Model} & \textbf{16} & \textbf{24} & \textbf{32} & \textbf{64} \\ 
\midrule

Model & SLNet-S & 93.12 & 93.32 & \textbf{93.64} & 93.08 \\ 
Net40 & SLNet-M & 93.34 & 93.48 & \textbf{93.92} & 93.17 \\ 

\midrule

Scan & SLNet-S & 81.99 & \textbf{83.45} & 83.38 & 82.13 \\ 
Object & SLNet-M & 82.86 & \textbf{84.25} & 83.58 & 81.83 \\ 

\bottomrule

\end{tabular}
\end{adjustbox}
\end{table}

\noindent\textbf{Sampling strategy.}
FPS consistently outperforms random sampling (Table~\ref{tab:ablation_sampling}) by 0.3\,pp on ModelNet40 and 0.6\,pp on ScanObjectNN, the latter gap reflecting FPS's advantage in maintaining spatial coverage under occlusion and clutter.

\noindent\textbf{SLNet-T encoder, loss, and training design.}
Table~\ref{tab:ablation_s3dis} reports 10 key runs across three groups.

\textit{Encoder architecture (§A).}
Replacing the MLP encoder with full local Point Transformer attention yields a $+$9.6\,pp mIoU gain (43.68\,$\to$\,53.24\%), with the hybrid variant (attention only in the two coarsest stages) providing a midpoint at 48.79\%.
This large gap confirms that dense attention over local neighborhoods is essential for fine-grained scene understanding—shared MLPs lack the relational capacity to disambiguate adjacent categories such as \textit{wall}, \textit{column}, and \textit{beam} in cluttered indoor scenes.

\textit{Loss function (§B).}
Switching from plain cross-entropy (B1, 55.47\%) to inverse-square-root weighted CE (B2, 58.16\%) yields $+$2.7\,pp mIoU and $+$3.8\,pp mAcc, directly addressing S3DIS's severe class imbalance.
Focal loss ($\gamma\!=\!2$, B3, 55.84\%) is a closer competitor but still underperforms WCE, suggesting that the static frequency-based reweighting is more stable than adaptive down-weighting for this distribution.

\textit{Training design (§C).}
EMA is the most impactful regulariser ($-$0.9\,pp when removed; C1), while label smoothing ($\varepsilon\!=\!0.1$) provides a marginal 0.1\,pp benefit (C2).
Halving neighbourhood size to $k\!=\![8,8,16,32]$ cuts FLOPs by 44\% but costs 3.1\,pp mIoU (C3), a poor trade-off at full model scale.
Halving channel width (C4) reduces parameters by 75\% (2.46\,$\to$\,0.62\,M) and FLOPs by 75\% (6.49\,$\to$\,1.64\,G) at a cost of 5.5\,pp mIoU, an attractive operating point for extremely constrained deployments.

Twelve additional post-optimisation experiments (§D–G: loss refinements, rare-class targeting, extended training, and LR/EMA schedules) consistently confirmed B2 as the robust global optimum.
The closest challenger, G2 (EMA $\rho\!=\!0.9999$, 57.54\%), fell 0.62\,pp short; combining OneCycle LR with slow EMA (G3) caused training divergence due to the mismatch between the high-LR exploration phase and the slow-moving EMA checkpoint.

\begin{figure}[t]
  \centering
  \resizebox{0.95\linewidth}{!}{%
    \setlength{\tabcolsep}{1pt}
    \begin{tabular}{ccccc}

      \raisebox{0.5\height}{\rotatebox{90}{\tiny \textbf{DGCNN}}} &
      \includegraphics[width=0.11\textwidth]{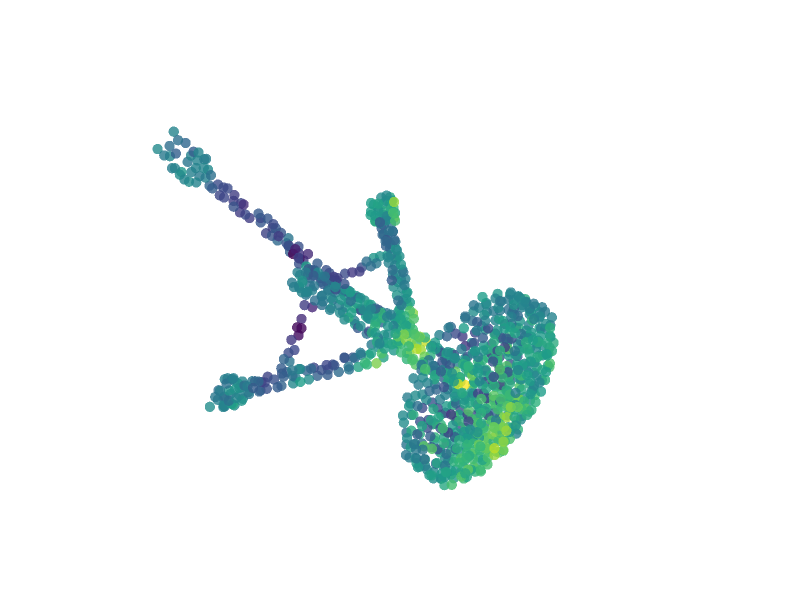} &
      \includegraphics[width=0.11\textwidth]{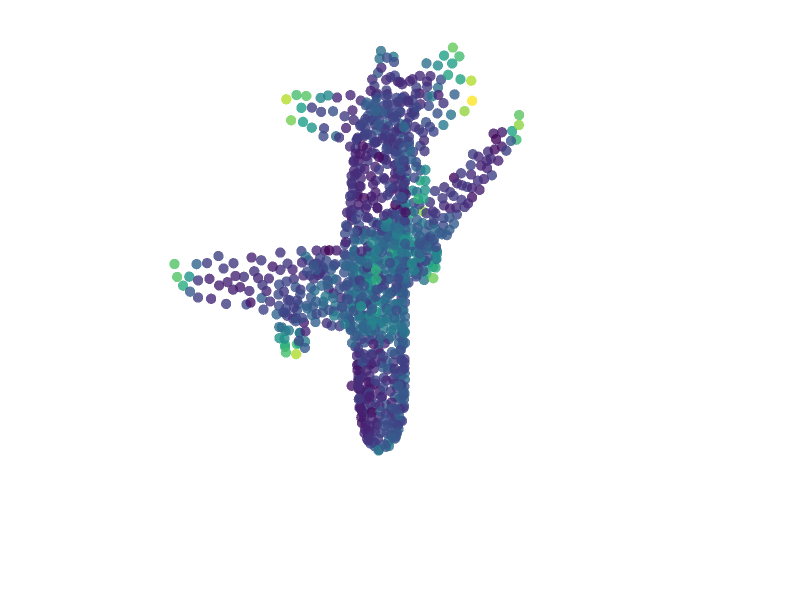} &
      \includegraphics[width=0.11\textwidth]{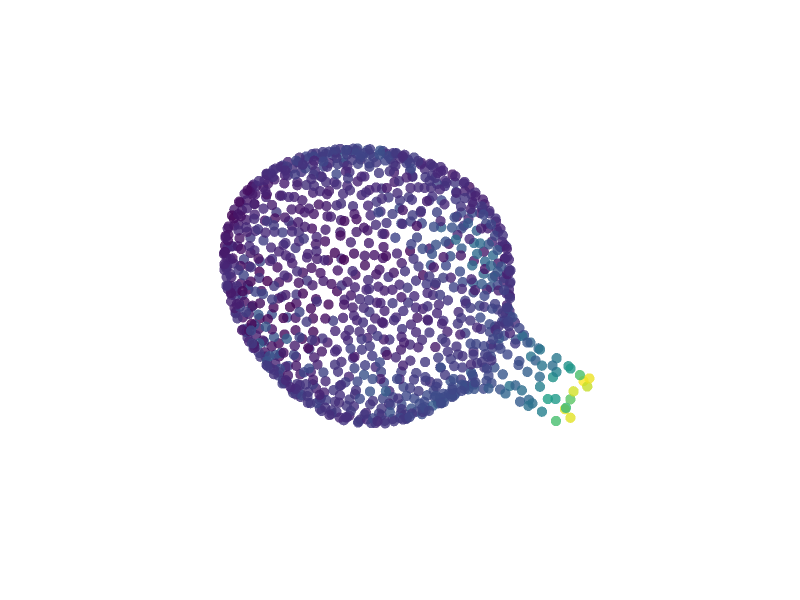} &
      \includegraphics[width=0.11\textwidth]{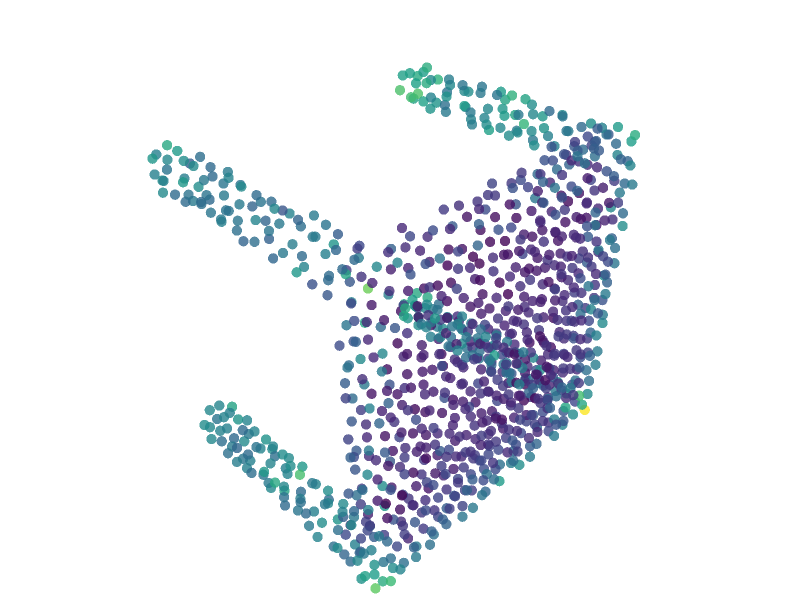} 
      \\
      
      \raisebox{0.5\height}{\rotatebox{90}{\tiny \textbf{NAPE 16}}} &
      \includegraphics[width=0.11\textwidth]{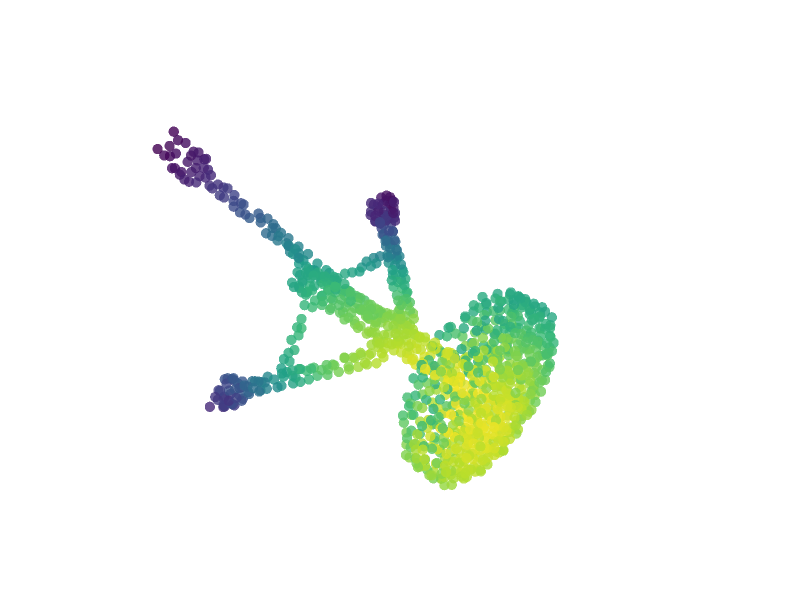} &
      \includegraphics[width=0.11\textwidth]{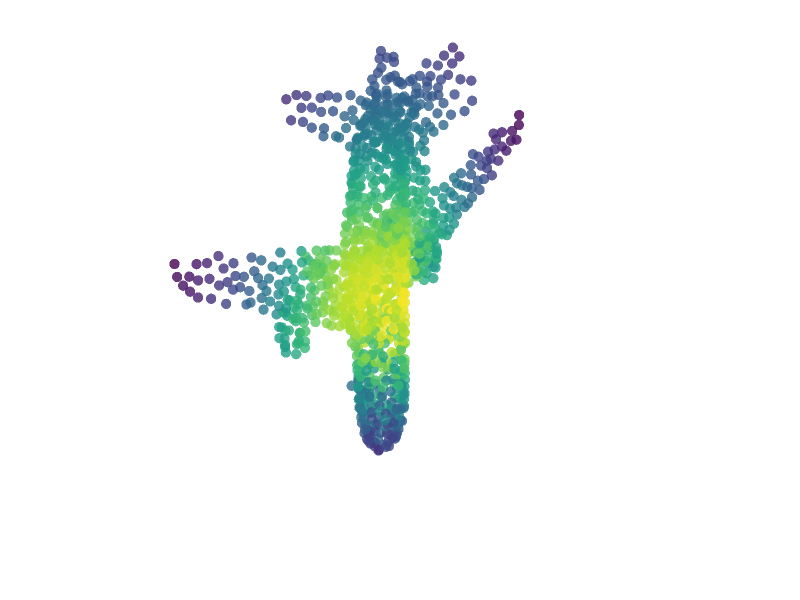} &
      \includegraphics[width=0.11\textwidth]{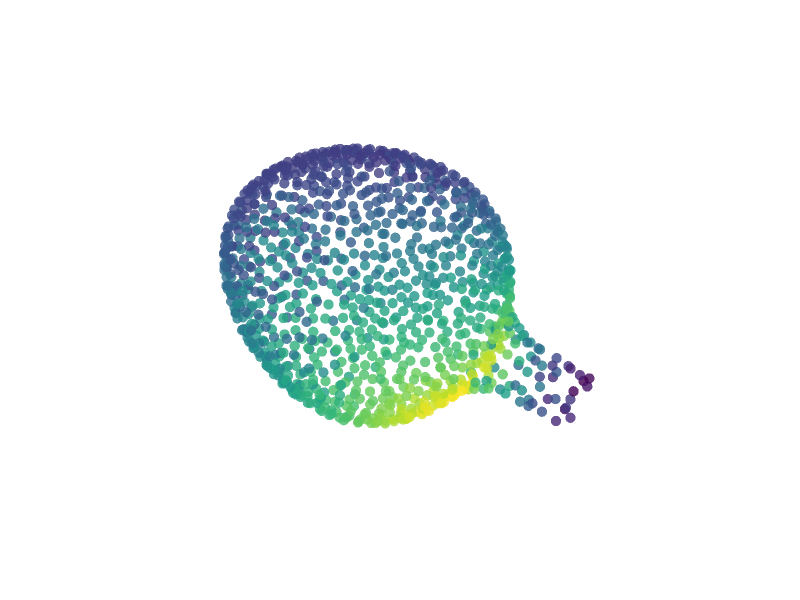} &
      \includegraphics[width=0.11\textwidth]{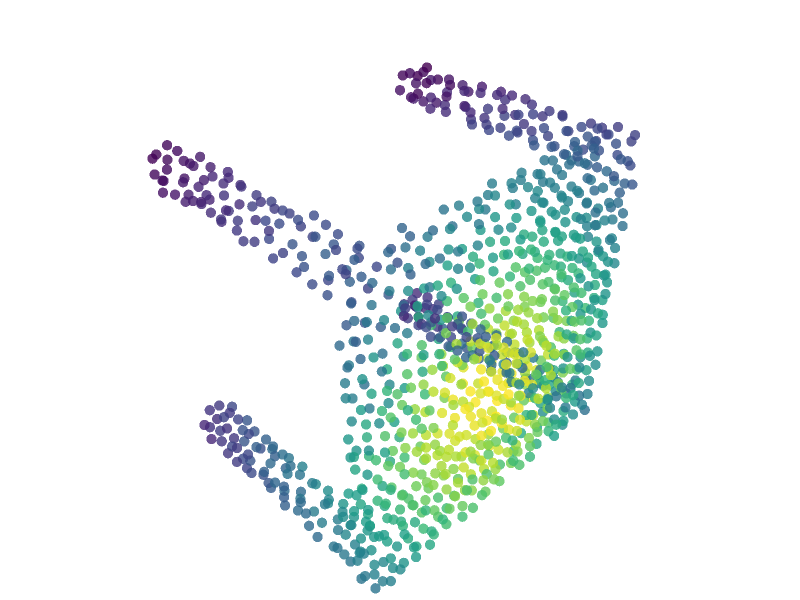} 
      \\

      \raisebox{0.5\height}{\rotatebox{90}{\tiny \textbf{NAPE 32}}} &
      \includegraphics[width=0.11\textwidth]{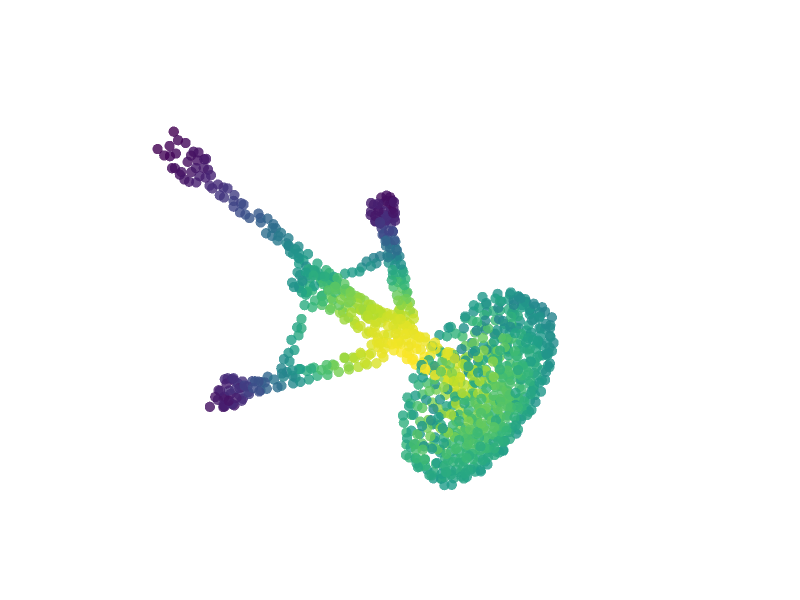} &
      \includegraphics[width=0.11\textwidth]{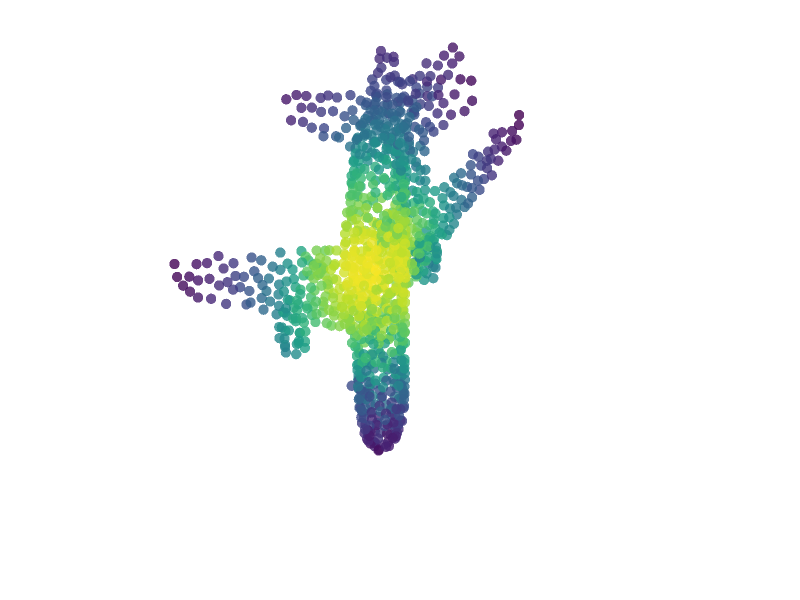} &
      \includegraphics[width=0.11\textwidth]{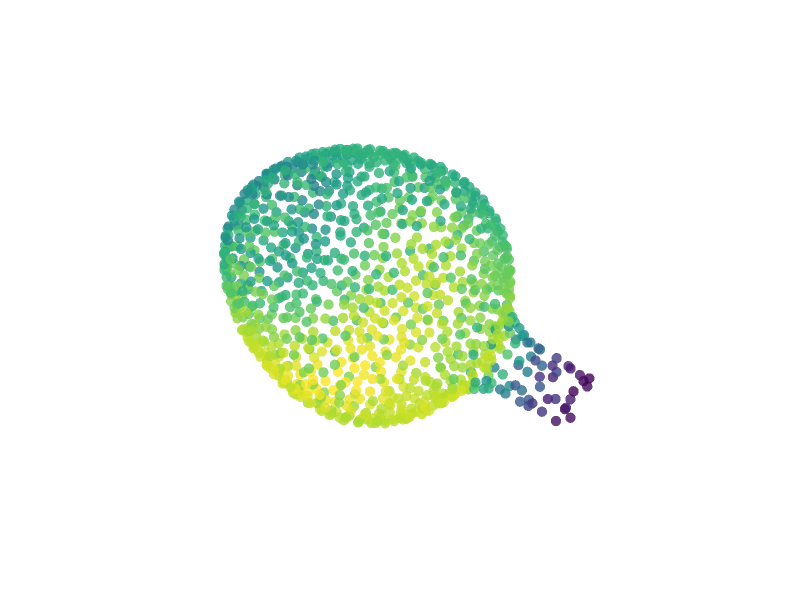} &
      \includegraphics[width=0.11\textwidth]{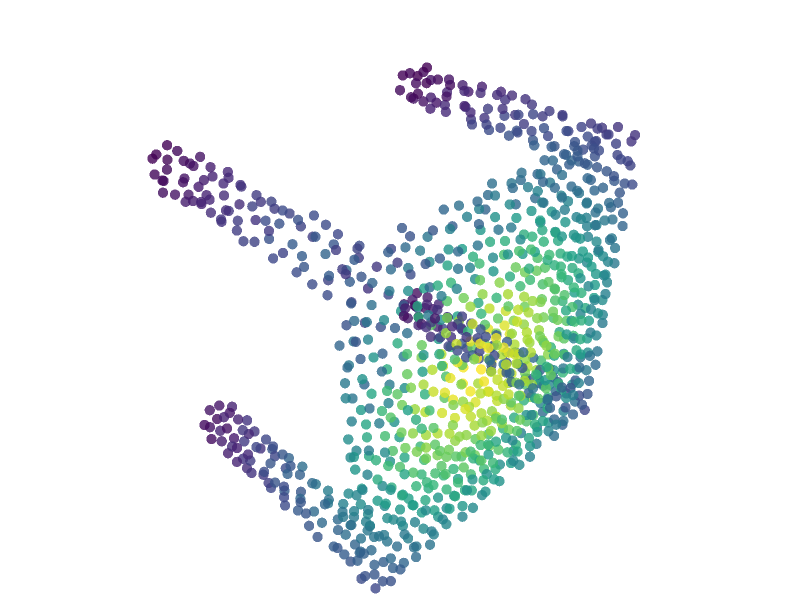} 
      \\

    \end{tabular}%
  }
  \caption{Qualitative comparison of saliency on ModelNet40 samples: top row—DGCNN stage-2 EdgeConv gradient saliency; middle and bottom—NAPE gradient saliency (channel widths 16 and 32). NAPE yields sharper, semantically focused responses despite operating at the input layer.}
  \label{fig:attention_maps}
\end{figure}

\paragraph{Qualitative.}
Gradient-based saliency maps from the NAPE block show more localized responses on semantic parts (e.g., chair legs and lamp shades), whereas DGCNN features appear more diffuse (Figure~\ref{fig:attention_maps}). This qualitative behavior is consistent with the quantitative trends observed in our experiments.
This behavior is consistent across channel widths (16 and 32) and the maps remain robust to small input perturbations, indicating stable geometric reasoning at early layers.

\begin{table}[t!]
\centering
\scriptsize



\caption{LRB bottleneck ratio ablation on ModelNet40 and ScanObjectNN (OA\,\%, params\,M). $\div 4$ is the Pareto-optimal point.}

\label{tab:ablation_residual}
\begin{adjustbox}{max width=\columnwidth}
\begin{tabular}{l|c|c|cccc}

\toprule
\textbf{Dataset} & \textbf{Model} & & \textbf{$\div 8$} & \textbf{$\div 4$} & \textbf{$\div 2$} & \textbf{$1$} \\
\midrule

\multirow{4}{*}{ModelNet40}
& \multirow{2}{*}{SLNet-S}
& Acc & 92.07 & \textbf{93.64} & 93.51 & 93.49 \\
& & Param & 0.11 & 0.14 & 0.20 & 0.31 \\
\cmidrule{2-7}
& \multirow{2}{*}{SLNet-M}
& Acc & 92.34 & \textbf{93.92} & 93.79 & 93.65 \\
& & Param & 0.44 & 0.55 & 0.77 & 1.20 \\

\midrule

\multirow{4}{*}{ScanObjecNN}
& \multirow{2}{*}{SLNet-S}
& Acc & 81.23 & \textbf{83.45} & 83.28 & 83.00 \\
& & Param & 0.09 & 0.12 & 0.19 & 0.31 \\
\cmidrule{2-7}
& \multirow{2}{*}{SLNet-M}
& Acc & 82.72 & \textbf{84.25} & 84.11 & 83.82 \\
& & Param & 0.35 & 0.48 & 0.73 & 1.23 \\

\bottomrule

\end{tabular}
\end{adjustbox}
\end{table}

\begin{table}[b!]
\centering
\scriptsize

\caption{FPS vs.\ random sampling on ModelNet40 and ScanObjectNN (OA\,\%). Larger gains on real-world data.}
\label{tab:ablation_sampling}

\begin{adjustbox}{max width=\columnwidth}
\begin{tabular}{c|c|cc}

\toprule
\textbf{Dataset} & \textbf{Model} & \textbf{Random Sampling} & \textbf{FPS} \\

\midrule

\multirow{2}{*}{ModelNet40}
& SLNet-S & 93.31 & \textbf{93.64} \\
& SLNet-M & 93.68 & \textbf{93.92} \\

\midrule

\multirow{2}{*}{ScanObjectNN}
& SLNet-S & 82.83 & \textbf{83.45} \\
& SLNet-M & 83.86 & \textbf{84.25} \\

\bottomrule

\end{tabular}
\end{adjustbox}
\end{table}

\begin{table}[!b]
\centering
\setlength{\tabcolsep}{3.5pt}
\renewcommand{\arraystretch}{1.08}
\scriptsize
\caption{%
  SLNet-T ablation on S3DIS Area\,5.
  Three groups: encoder architecture~(§A), loss function~(§B), and
  training-design components~(§C).
  Each group's fixed settings and abbreviations follow note\,$^{a}$.
  \textbf{Bold}: best mIoU per group.
  ${\boldsymbol\star}$: global best (B2).
  ${\dagger}$: proposed encoder baseline.
}
\label{tab:ablation_s3dis}
\begin{adjustbox}{max width=\columnwidth}
\begin{tabular}{@{}l p{2.5cm} c l  r r r  r r@{}}
\toprule
\multicolumn{4}{@{}l}{\textit{Configuration}} &
\multicolumn{3}{c}{\textit{Accuracy\,(\%)}} &
\multicolumn{2}{c@{}}{\textit{Cost}} \\
\cmidrule(r){1-4}\cmidrule(lr){5-7}\cmidrule(l){8-9}
{ID} & {Description} & {Enc} & {Loss} &
{mIoU} & {OA} & {mAcc} &
{Param\,(M)} & {FLOPs\,(G)} \\
\midrule
%
\rowcolor{gray!12}
\multicolumn{9}{@{}l}{\rule{0pt}{7pt}\textbf{§A}\enspace\textit{Encoder Architecture}\enspace{\scriptsize(loss: F+L)}}\\[-1pt]
A1 & MLP aggregation        & M & F+L & 43.68 & 77.88 & 51.84 & 1.16 & 2.74 \\
A2 & Hybrid attn (st.\,3--4)& H & F+L & 48.79 & 80.10 & 56.02 & 2.39 & 5.60 \\
A3$^{\dagger}$ & Full attention & A & F+L & \textbf{53.24} & 84.08 & 60.03 & 2.46 & 6.49 \\
\midrule
%
\rowcolor{gray!12}
\multicolumn{9}{@{}l}{\rule{0pt}{7pt}\textbf{§B}\enspace\textit{Loss Function}\enspace{\scriptsize(enc: A;\enspace base: A3$^{\dagger}$,\,53.24\,\%)}}\\[-1pt]
B1 & Plain CE, $\varepsilon\!=\!0$            & A & CE  & 55.47 & 85.19 & 61.63 & 2.46 & 6.49 \\
\rowcolor{lightgreen}
B2$^{\star}$ & Weighted CE ($1/\!\sqrt{f_c}$) & A & WCE & \textbf{58.16} & 85.30 & 65.40 & 2.46 & 6.49 \\
B3 & Focal, $\gamma\!=\!2$                    & A & Foc & 55.84 & 85.28 & 61.93 & 2.46 & 6.49 \\
\midrule
%
\rowcolor{gray!12}
\multicolumn{9}{@{}l}{\rule{0pt}{7pt}\textbf{§C}\enspace\textit{Training Design}\enspace{\scriptsize(enc: A,\;loss: F+L;\enspace base: A3$^{\dagger}$,\,53.24\,\%)}}\\[-1pt]
C1 & No EMA                      & A & F+L & 52.30 & 83.53 & 58.78 & 2.46 & 6.49 \\
C2 & No label smoothing          & A & F+L & \textbf{53.30} & 83.36 & 60.25 & 2.46 & 6.49 \\
C3 & $k\!=\![8,8,16,32]$         & A & F+L & 50.17 & 82.37 & 57.58 & 2.46 & 3.66 \\
C4 & Half-channel width$^{b}$    & A & F+L & 47.73 & 81.08 & 54.57 & \textbf{0.62} & \textbf{1.64} \\
\bottomrule
\end{tabular}
\end{adjustbox}
\vspace{3pt}
\begin{minipage}{\columnwidth}
\setlength{\parindent}{0pt}
\scriptsize
$^{a}$\,\textbf{Common settings:}
$N\!=\!16{,}384$\,pts/crop;\;
crop radius $r\!=\!2.5$\,m;\;
$k\!=\![16,16,32,64]$;\;
encoder dims $[64,128,256,512]$
(C4: $[32,64,128,256]$);\;
decoder dims $[256,128,64,64]$
(C4: $[128,64,32,32]$);\;
AdamW $\eta\!=\!9{\times}10^{-4}$, $\lambda\!=\!2{\times}10^{-3}$;\;
cosine LR, EMA $\rho\!=\!0.999$;\;
label smoothing $\varepsilon\!=\!0.1$ (C2: 0);\;
batch 48/16;\;seed 42.
\textbf{Enc:}\;M\,=\,MLP,\;H\,=\,Hybrid,\;A\,=\,Attn.
\textbf{Loss:}\;CE;\;WCE\,=\,weighted CE;\;Foc\,=\,focal;\;F+L\,=\,focal+Lov\'{a}sz.
$^{b}$\,C4 bold cost entries indicate best within §C on those metrics.
\end{minipage}
\end{table}
\section{Conclusion}
\label{sec:conclusion}

We presented \textbf{\SLNet}, a lightweight hierarchical backbone for 3D point cloud understanding built on two simple ideas: parameter-free geometric encoding and minimal learned modulation. \textbf{NAPE} captures raw XYZ geometry through an input-adaptive combination of Gaussian RBF and cosine bases without introducing learnable parameters, while \textbf{GMU} provides a lightweight per-channel affine recalibration with only $2D$ learnable scalars. Across object classification, few-shot learning, part segmentation, and scene segmentation, these design choices lead to a strong balance between recognition performance and deployment efficiency.

For object-level tasks, \SLNetS and \SLNetM achieve competitive accuracy with substantially fewer parameters and lower compute cost than many stronger supervised baselines. In few-shot classification, both variants also outperform non-parametric baselines in the 10-way 20-shot setting without pretraining. For large-scale indoor scene segmentation, \SLNetT extends the same design philosophy with local Point Transformer attention and reaches a favorable efficiency-accuracy trade-off under a much smaller model budget than prior transformer-based alternatives. We further introduced \textbf{NetScore$^{+}$}, which incorporates latency and memory alongside standard efficiency measures to better reflect deployment-oriented performance. Overall, our results suggest that compact point cloud models can remain highly competitive when geometric encoding, feature modulation, and local aggregation are designed explicitly for efficiency.


\bibliographystyle{IEEEtran}
\bibliography{ref}

@article{saeid2026npnet,
  title={NPNet: A Non-Parametric Network with Adaptive Gaussian-Fourier Positional Encoding for 3D Classification and Segmentation},
  author={Saeid, Mohammad and Salarpour, Amir and MohajerAnsari, Pedram and Pes{\'e}, Mert D},
  journal={arXiv preprint arXiv:2602.00542},
  year={2026}
}

@article{he2025pointdiffuse,
  title={PointDiffuse: a dual-conditional diffusion model for enhanced point cloud semantic segmentation},
  author={He, Yong and Yu, Hongshan and Feng, Mingtao and Chen, Tongjia and Li, Zechuan and Ulhaq, Anwaar and Anwar, Saeed and Mian, Ajmal Saeed},
  journal={arXiv preprint arXiv:2503.06094},
  year={2025}
}

@article{dumic2025three,
  title={Three-Dimensional Point Cloud Applications, Datasets, and Compression Methodologies for Remote Sensing: A Meta-Survey},
  author={Dumic, Emil and da Silva Cruz, Lu{\'\i}s A},
  journal={Sensors},
  volume={25},
  number={6},
  pages={1660},
  year={2025},
  publisher={MDPI}
}

@article{mohammadi2024point,
  title={Point-GN: A Non-Parametric Network Using Gaussian Positional Encoding for Point Cloud Classification},
  author={Mohammadi, Marzieh and Salarpour, Amir},
  journal={arXiv preprint arXiv:2412.03056},
  year={2024}
}

@inproceedings{zhang2020pointhop++,
  title={Pointhop++: A lightweight learning model on point sets for 3d classification},
  author={Zhang, Min and Wang, Yifan and Kadam, Pranav and Liu, Shan and Kuo, C-C Jay},
  booktitle={2020 IEEE International Conference on Image Processing (ICIP)},
  pages={3319--3323},
  year={2020},
  organization={IEEE}
}

@article{duan2021robotics,
  title={Robotics dexterous grasping: The methods based on point cloud and deep learning},
  author={Duan, Haonan and Wang, Peng and Huang, Yayu and Xu, Guangyun and Wei, Wei and Shen, Xiaofei},
  journal={Frontiers in Neurorobotics},
  volume={15},
  pages={658280},
  year={2021},
  publisher={Frontiers Media SA}
}

@inproceedings{zhao2021point,
  title={Point transformer},
  author={Zhao, Hengshuang and Jiang, Li and Jia, Jiaya and Torr, Philip HS and Koltun, Vladlen},
  booktitle={Proceedings of the IEEE/CVF international conference on computer vision},
  pages={16259--16268},
  year={2021}
}

@article{saeid2025enhancing,
  title={Enhancing 3D Point Cloud Classification with ModelNet-R and Point-SkipNet},
  author={Saeid, Mohammad and Salarpour, Amir and MohajerAnsari, Pedram},
  journal={arXiv preprint arXiv:2509.05198},
  year={2025}
}

@article{zhang2023parameter,
  title={Parameter is not all you need: Starting from non-parametric networks for 3d point cloud analysis},
  author={Zhang, Renrui and Wang, Liuhui and Guo, Ziyu and Wang, Yali and Gao, Peng and Li, Hongsheng and Shi, Jianbo},
  journal={arXiv preprint arXiv:2303.08134},
  year={2023}
}

@article{salarpour2024pointgn,
  title={Point-GN: A Non-Parametric Network Using Gaussian Positional Encoding for Point Cloud Classification},
  author={Mohammadi, Marzieh and Salarpour, Amir},
  journal={arXiv preprint arXiv:2412.03056},
  year={2024}
}

@article{salarpour2025pointln,
  author= {Marzieh Mohammadi and Amir Salarpour and Pedram MohajerAnsari},
  title= {Point-LN: A Lightweight Framework for Efficient Point Cloud Classification Using Non-Parametric Positional Encoding},
  journal   = {arXiv preprint arXiv:2501.14238},
  year      = {2025}
}

@article{mahmood2020bim,
  title={BIM-based registration and localization of 3D point clouds of indoor scenes using geometric features for augmented reality},
  author={Mahmood, Bilawal and Han, SangUk and Lee, Dong-Eun},
  journal={Remote Sensing},
  volume={12},
  number={14},
  pages={2302},
  year={2020},
  publisher={MDPI}
}

@article{guo2020deep,
  title={Deep learning for 3d point clouds: A survey},
  author={Guo, Yulan and Wang, Hanyun and Hu, Qingyong and Liu, Hao and Liu, Li and Bennamoun, Mohammed},
  journal={IEEE transactions on pattern analysis and machine intelligence},
  volume={43},
  number={12},
  pages={4338--4364},
  year={2020},
  publisher={IEEE}
}

@article{li2020deep,
  title={Deep learning for lidar point clouds in autonomous driving: A review},
  author={Li, Ying and Ma, Lingfei and Zhong, Zilong and Liu, Fei and Chapman, Michael A and Cao, Dongpu and Li, Jonathan},
  journal={IEEE Transactions on Neural Networks and Learning Systems},
  volume={32},
  number={8},
  pages={3412--3432},
  year={2020},
  publisher={IEEE}
}

@article{atzmon2018point,
  title={Point convolutional neural networks by extension operators},
  author={Atzmon, Matan and Maron, Haggai and Lipman, Yaron},
  journal={arXiv preprint arXiv:1803.10091},
  year={2018}
}

@article{wong2018netscore,
  title={NetScore: Towards Universal Metrics for Large-scale Performance Analysis of Deep Neural Networks for Practical Usage},
  author={Wong, Alexander},
  journal={arXiv preprint arXiv:1806.05512},
  year={2018}
}

@article{sharma2020self,
  title={Self-supervised few-shot learning on point clouds},
  author={Sharma, Charu and Kaul, Manohar},
  journal={Advances in Neural Information Processing Systems},
  volume={33},
  pages={7212--7221},
  year={2020}
}

@inproceedings{yu2022point,
  title={Point-bert: Pre-training 3d point cloud transformers with masked point modeling},
  author={Yu, Xumin and Tang, Lulu and Rao, Yongming and Huang, Tiejun and Zhou, Jie and Lu, Jiwen},
  booktitle={Proceedings of the IEEE/CVF conference on computer vision and pattern recognition},
  pages={19313--19322},
  year={2022}
}

@inproceedings{pang2022masked,
  title={Masked autoencoders for point cloud self-supervised learning},
  author={Pang, Yatian and Wang, Wenxiao and Tay, Francis EH and Liu, Wei and Tian, Yonghong and Yuan, Li},
  booktitle={European conference on computer vision},
  pages={604--621},
  year={2022},
  organization={Springer}
}

@inproceedings{qi2017pointnet,
  title={Pointnet: Deep learning on point sets for 3d classification and segmentation},
  author={Qi, Charles R and Su, Hao and Mo, Kaichun and Guibas, Leonidas J},
  booktitle={Proceedings of the IEEE conference on computer vision and pattern recognition},
  pages={652--660},
  year={2017}
}

@article{qi2017pointnet++,
  title={Pointnet++: Deep hierarchical feature learning on point sets in a metric space},
  author={Qi, Charles Ruizhongtai and Yi, Li and Su, Hao and Guibas, Leonidas J},
  journal={Advances in neural information processing systems},
  volume={30},
  year={2017}
}

@article{li2018pointcnn,
  title={Pointcnn: Convolution on x-transformed points},
  author={Li, Yangyan and Bu, Rui and Sun, Mingchao and Wu, Wei and Di, Xinhan and Chen, Baoquan},
  journal={Advances in neural information processing systems},
  volume={31},
  year={2018}
}

@inproceedings{wu2019pointconv,
  title={Pointconv: Deep convolutional networks on 3d point clouds},
  author={Wu, Wenxuan and Qi, Zhongang and Fuxin, Li},
  booktitle={Proceedings of the IEEE/CVF Conference on computer vision and pattern recognition},
  pages={9621--9630},
  year={2019}
}

@inproceedings{thomas2019kpconv,
  title={Kpconv: Flexible and deformable convolution for point clouds},
  author={Thomas, Hugues and Qi, Charles R and Deschaud, Jean-Emmanuel and Marcotegui, Beatriz and Goulette, Fran{\c{c}}ois and Guibas, Leonidas J},
  booktitle={Proceedings of the IEEE/CVF international conference on computer vision},
  pages={6411--6420},
  year={2019}
}

@article{wang2019dynamic,
  title={Dynamic graph cnn for learning on point clouds},
  author={Wang, Yue and Sun, Yongbin and Liu, Ziwei and Sarma, Sanjay E and Bronstein, Michael M and Solomon, Justin M},
  journal={ACM Transactions on Graphics (tog)},
  volume={38},
  number={5},
  pages={1--12},
  year={2019},
  publisher={Acm New York, NY, USA}
}

@article{guo2021pct,
  title={Pct: Point cloud transformer},
  author={Guo, Meng-Hao and Cai, Jun-Xiong and Liu, Zheng-Ning and Mu, Tai-Jiang and Martin, Ralph R and Hu, Shi-Min},
  journal={Computational visual media},
  volume={7},
  pages={187--199},
  year={2021},
  publisher={Springer}
}

@inproceedings{wu2023attention,
  title={Attention-based point cloud edge sampling},
  author={Wu, Chengzhi and Zheng, Junwei and Pfrommer, Julius and Beyerer, J{\"u}rgen},
  booktitle={Proceedings of the IEEE/CVF Conference on Computer Vision and Pattern Recognition},
  pages={5333--5343},
  year={2023}
}

@inproceedings{xiang2021walk,
  title={Walk in the cloud: Learning curves for point clouds shape analysis},
  author={Xiang, Tiange and Zhang, Chaoyi and Song, Yang and Yu, Jianhui and Cai, Weidong},
  booktitle={Proceedings of the IEEE/CVF international conference on computer vision},
  pages={915--924},
  year={2021}
}

@article{ma2022rethinking,
  title={Rethinking network design and local geometry in point cloud: A simple residual MLP framework},
  author={Ma, Xu and Qin, Can and You, Haoxuan and Ran, Haoxi and Fu, Yun},
  journal={arXiv preprint arXiv:2202.07123},
  year={2022}
}

@inproceedings{zhou2024dynamic,
  title={Dynamic adapter meets prompt tuning: Parameter-efficient transfer learning for point cloud analysis},
  author={Zhou, Xin and Liang, Dingkang and Xu, Wei and Zhu, Xingkui and Xu, Yihan and Zou, Zhikang and Bai, Xiang},
  booktitle={Proceedings of the IEEE/CVF Conference on Computer Vision and Pattern Recognition},
  pages={14707--14717},
  year={2024}
}

@inproceedings{yang2018foldingnet,
  title={Foldingnet: Point cloud auto-encoder via deep grid deformation},
  author={Yang, Yaoqing and Feng, Chen and Shen, Yiru and Tian, Dong},
  booktitle={Proceedings of the IEEE conference on computer vision and pattern recognition},
  pages={206--215},
  year={2018}
}

@article{wu2016learning,
  title={Learning a probabilistic latent space of object shapes via 3d generative-adversarial modeling},
  author={Wu, Jiajun and Zhang, Chengkai and Xue, Tianfan and Freeman, Bill and Tenenbaum, Josh},
  journal={Advances in neural information processing systems},
  volume={29},
  year={2016}
}

@inproceedings{sun2024parameter,
  title={Parameter-efficient prompt learning for 3d point cloud understanding},
  author={Sun, Hongyu and Wang, Yongcai and Chen, Wang and Deng, Haoran and Li, Deying},
  booktitle={2024 IEEE International Conference on Robotics and Automation (ICRA)},
  pages={9478--9486},
  year={2024},
  organization={IEEE}
}

@inproceedings{wang2021fast,
  title={Fast k-nn graph construction by gpu based nn-descent},
  author={Wang, Hui and Zhao, Wan-Lei and Zeng, Xiangxiang and Yang, Jianye},
  booktitle={Proceedings of the 30th ACM International Conference on Information \& Knowledge Management},
  pages={1929--1938},
  year={2021}
}

@article{hou2024graph,
  title={Graph Construction with Flexible Nodes for Traffic Demand Prediction},
  author={Hou, Jinyan and Liu, Shan and Zhang, Ya and Qin, Haotong},
  journal={arXiv preprint arXiv:2403.00276},
  year={2024}
}

@article{su2020dv,
  title={DV-ConvNet: Fully convolutional deep learning on point clouds with dynamic voxelization and 3d group convolution},
  author={Su, Zhaoyu and Tan, Pin Siang and Chow, Junkang and Wu, Jimmy and Cheong, Yehur and Wang, Yu-Hsing},
  journal={arXiv preprint arXiv:2009.02918},
  year={2020}
}

@inproceedings{wu2024point,
  title={Point transformer v3: Simpler faster stronger},
  author={Wu, Xiaoyang and Jiang, Li and Wang, Peng-Shuai and Liu, Zhijian and Liu, Xihui and Qiao, Yu and Ouyang, Wanli and He, Tong and Zhao, Hengshuang},
  booktitle={Proceedings of the IEEE/CVF conference on computer vision and pattern recognition},
  pages={4840--4851},
  year={2024}
}

\end{document}